\documentclass[preprint,authoryear,12pt]{elsarticle}
\makeatletter
\def\ps@pprintTitle{%
  \let\@oddhead\@empty
  \let\@evenhead\@empty
  \let\@oddfoot\@empty
  \let\@evenfoot\@empty}
\makeatother

\usepackage{amssymb}
\usepackage{amsmath}
\usepackage{hyperref}
\usepackage{graphicx}
\usepackage{subcaption}
\usepackage{caption}
\usepackage{multirow}
\usepackage{rotating}
\usepackage{comment}
\usepackage[table]{xcolor}

\usepackage{booktabs}
\usepackage{array}
\usepackage{pifont}
\usepackage{xspace}
\usepackage{float}
\usepackage[authoryear]{natbib}
\usepackage{lineno}
\usepackage{makecell}

\newcommand{\One}{%
  \mbox{\raisebox{-.2ex}{\includegraphics[width=2ex]{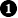}}}%
  \hspace{0.3em}%
  {\textsf{satellite images only}}\xspace%
}

\newcommand{\Two}{%
  \mbox{\raisebox{-.2ex}{\includegraphics[width=2ex]{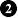}}}%
  \hspace{0.3em}%
  {\textsf{environmental predictors only}}\xspace%
}

\newcommand{\Three}{%
  \mbox{\raisebox{-.2ex}{\includegraphics[width=2ex]{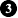}}}%
  \hspace{0.3em}%
  {\textsf{satellite and environmental data}}\xspace%
}

\newcommand{\cmark}{\ding{51}} 
\newcommand{\xmark}{\ding{55}} 

\journal{ISPRS}

\begin{document}

\begin{frontmatter}



\title{CanadaFireSat: Toward high-resolution wildfire forecasting with multiple modalities}

\author[inst1]{Hugo Porta\corref{cor1}} 
\author[inst2]{Emanuele Dalsasso}
\author[inst3]{Jessica L. McCarty}
\author[inst1]{Devis Tuia}

\cortext[cor1]{Corresponding author: hugo.porta@epfl.ch, Published in \textit{ISPRS Journal of Photogrammetry and Remote Sensing}. DOI: \href{https://doi.org/10.1016/j.isprsjprs.2026.05.050}{10.1016/j.isprsjprs.2026.05.050}.}

\affiliation[inst1]{organization={EPFL},
            addressline={Route des Ronquos 86}, 
            city={Sion},
            postcode={1950}, 
            state={Wallis},
            country={Switzerland}}

\affiliation[inst2]{o={Université Grenoble Alpes, Inria, CNRS, Grenoble INP, LJK}, c={Grenoble}, p={38000}, cy={France}}

\affiliation[inst3]{organization={NASA Ames Research Center, Earth Science Division,},
            addressline={Moffett Field}, 
            state={California},
            postcode={94035},
            country={USA}}

\begin{abstract}

Canada experienced in 2023 one of the most severe wildfire seasons in recent history, causing damage across ecosystems, destroying communities, and emitting large quantities of $\text{CO}_2$. This extreme wildfire season is symptomatic of a climate-change-induced increase in length and severity of fire seasons affecting the boreal ecosystem. Therefore, it is critical to empower wildfire management in boreal communities with better monitoring solutions. Wildfire probability maps are an important tool for understanding the likelihood of wildfire occurrence and the potential severity of future wildfires. 
Fire forecasting tools based on Earth observation data exist, but they are limited both by the lack of label information and by their reliance on coarse-resolution environmental drivers and satellite products, which leads to wildfire occurrence prediction of reduced resolution, typically around $\sim 0.1$°. To tackle these two limitations, this paper presents a benchmark dataset, CanadaFireSat available on the \href{https://huggingface.co/datasets/EPFL-ECEO/CanadaFireSat}{Hugging Face Hub}, and baseline methods for high-resolution wildfire forecasting. We model wildfire forecasting as a binary patch classification task at $100$ m, where each patch is labeled as \textit{fire} if at least one pixel within the patch has burned at the native label resolution ($10$ m). CanadaFireSat leverages multi-modal data from high-resolution multi-spectral satellite images (Sentinel-2), mid-resolution satellite products (MODIS), and environmental factors (ERA5). 
We experiment with convolutional (CNN) and transformer (ViT) architectures. We observe that using multi-modal temporal inputs outperforms single-modal temporal inputs across all metrics, achieving a peak performance of $60.3 \%$ in F1 score for the 2023 wildfire season, a season never seen during model training. This demonstrates the potential of multi-modal deep learning for wildfire forecasting at high-resolution and continental scale. The code is available on GitHub for the \href{https://github.com/eceo-epfl/CanadaFireSat-Data}{data generation} and the \href{https://github.com/eceo-epfl/CanadaFireSat-Model}{model benchmarking}.
\end{abstract}


\end{frontmatter}



\section{Introduction}
\label{introduction}

As climate change accelerates, forests represent a key ecosystem to protect as they act as one of the main terrestrial carbon sinks \citep{keenan2018terrestrial}, a shelter for a major part of Earth's biodiversity \citep{lindenmayer2013conserving} 
, and a critical environment for numerous fragile human communities \citep{fernandez2021scientists}. 
In particular, the boreal ecosystem is a subarctic biome in the high northern latitudes characterized by coniferous and mixed deciduous-coniferous forests. They represent one of the largest terrestrial carbon sinks, with approximately $367.3$ petagrams to $1715.8$ petagrams of carbon stored \citep{bradshaw2015global}. However, they are at risk of permafrost thaw due to land impacts \citep{li2021influences} and are increasingly subject to long and devastating wildfire seasons \citep{mccarty2021reviews}. 

While wildfires severity is amplifying globally (areas burned by forest fires have seen a steady yearly increase of $\sim 5 \%$ since 2001 \citep{tyukavina2022global}), its effect is particularly devastating for the boreal ecosystem, representing roughly $70 \%$ of the fire-related tree cover loss \citep{tyukavina2022global} and where single wildfire events, like those during the 2023 Canadian wildfires season, can compete with annual $\text{CO}_2$ emissions of major industrialized nations \citep{byrne2024carbon}. 
Locally, boreal wildfires have a direct impact on the land surface as they directly increase permafrost thaw \citep{li2021influences, zhao2024forest}, and contribute to vegetation shifts to more fire-prone grassland-/steppe-dominant landscapes, as well as dry peat \citep{zhao2024forest, mccarty2021reviews}. Boreal forests span $58 \%$ of Canada's land mass. Therefore, in this paper, we use Canada as the area of interest to tackle the problem of wildfire forecasting in boreal ecosystems.

The use of remote sensing data to map and monitor wildfires has expanded, with studies considering satellite-based observations of vegetative fuel conditions, individual fire events, and the impacts of smoke. Numerous wildfire tools exist, with three main use cases focused on the different phases of a wildfire: before a wildfire occurs (pre-fire), during a wildfire (active fire), and after (post-fire). For active fires, satellite products and models detect active fire "hotspots" in near real-time \citep{de2021active} even up to 10-minutes via high-frequency geostationary observations \citep{zhang2023beyond, zhang202510} or predict wildfire spread via supervised prediction models \citep{huot2022next} and stochastic generative forecasting models \citep{hoang2022wildfire}, providing tools for wildfire management and decision-making. In the aftermath of wildfires (post-fire), methods to segment burned areas include supervised \citep{hu2023large} and unsupervised deep learning models \citep{zhang2022unsupervised}, enabling the evaluation of wildfire emissions, from $\text{CO}_2$ to black carbon, and the estimation of the impact of wildfires on the local ecosystem, natural resources, and communities. This paper focuses on the pre-fire phase, shaped from a methodological perspective as a forecasting task. Wildfire forecasting, often referred to as wildfire susceptibility or likelihood modeling \citep{pelletier2023wildfire, zhang2021deep}, aims to predict the spatial probability of a wildfire occurring in a given time horizon and at a given spatial resolution. This is done by producing wildfire probability maps. Wildfire forecasting is particularly useful for wildfire management by supporting staff and resource planning \citep{wotton2009interpreting}.

Wildfire forecasting is, by definition, a difficult task since it seeks to represent a complex and stochastic phenomenon. 
Fire susceptibility depends on several drivers: \textbf{i)} hydrometeorological conditions are the main variables that impact the suitability of vegetative fuels for combustion (ie, 'dryness') and fire propagation \citep{krawchuk2009global}. Fire susceptibility is also directly linked to \textbf{ii)} the available biomass for combustion, which depends on \textbf{iii)} the type of vegetation and other indicators such as \textbf{iv)} dead or live fuel moisture \citep{krawchuk2009global}. Climate change makes those predictors for wildfire occurrence non-stationary. For example, meteorological patterns are highly variable in the boreal biome, which can lead to extreme fire seasons \citep{mccarty2021reviews}. In addition, numerous vegetation changes have been observed or predicted, such as permafrost thaw \citep{li2021influences, zhao2024forest}, peatland destruction \citep{bourgeau2022assessing}, or a shift from coniferous to deciduous forest \citep{mccarty2021reviews}. Moreover, for effective wildfire forecasting, it is necessary to estimate \textbf{v)} the probability of ignition caused by humans or lightning \citep{perez2023variation} through proxies such as the proximity to human settlements. This variability implies that for similar environmental conditions, a wildfire may or may not occur based on latent variables for the model, which makes wildfire forecasting an especially complex task where ignition, even from lightning-only, is hard to model via traditional machine learning models \citep{coughlan2021using} and Bayesian probabilistic frameworks \citep{bates2021bayesian}. 

Historically, wildfire forecasting was performed by producing fire weather indices, such as the Canadian Forest Fire Weather Index (FWI) or the National Fire Danger Rating System (NFDRS), which are mainly driven by meteorological and fuel moisture data. Fire weather indices aim to represent complex relationships between fire predictors through constrained equations based on simplifying assumptions, such as the forest type (e.g "Pinus Banksiana") and neglecting the topography, leading to necessary recalibrations of the indices for specific areas like Switzerland \citep{steinfeld2022assessing} and the UK \citep{de2016calibration}. 
For instance, across Canada, the FWI has limitations in properly identifying the hydrometeorological conditions for combustion across all land cover types, and particularly so in peatlands \citep{waddington2012examining}. Moreover, those indices cannot approximate the stochastic character of wildfire occurrence as they focus on flammability conditions. In parallel, traditional machine learning (ML) algorithms based on handcrafted features were proposed in \citep{martell1987logistic, martell1989modelling} to identify drivers linked to wildfire occurrence, such as hydrometeorological conditions and human activities. These ML algorithms are limited in their ability to represent complex predictor relationships and possible spatio-temporal patterns, mostly because of the rigidity of the features used. Nevertheless, these methods are still widely utilized \citep{buch2023smlfire1, rodrigues2022firo} even in remote sensing \citep{maffei2021combining, chowdhury2015operational}.

\begin{table}[t]
\centering
\renewcommand{\arraystretch}{1.2}
\resizebox{\textwidth}{!}{%
\begin{tabular}{>{\raggedright\arraybackslash}p{3cm}
                >{\raggedright\arraybackslash}p{2.5cm}
                >{\raggedright\arraybackslash}p{2.5cm}
                >{\raggedright\arraybackslash}p{2.5cm}
                >{\raggedright\arraybackslash}p{2.5cm}
                >{\raggedright\arraybackslash}p{5cm}}
\toprule
\textbf{Predictor Type} & \textbf{Spatial Scale} & \textbf{Spatial Resolution} & \textbf{Temporal Resolution} & \textbf{Method Type} & \textbf{Models} \\
\midrule
& & & & Pixel-wise
& RF \citep{bakke2023data}, MLP \citep{joshi2021improving}, MDN \citep{buch2023smlfire1}, Hybrid ANFIS \citep{jaafari2019hybrid} \\
& & & & Spatial
& U-Net, ViT, TeleViT \citep{prapas2023televit}, CNN \citep{prapas2021deep}\\
& & & & Temporal
& LSTM \citep{natekar2021forest}\\
\multirow{-4}{*}{\makecell[l]{Environmental\\(+ Human)}}
& \multirow{-4}{*}{\makecell[l]{Local --\\Global}}
& \multirow{-4}{*}{\makecell[l]{30 m -- 1°}}
& \multirow{-4}{*}{\makecell[l]{Daily -- Static}}
& Spatio-Temporal
& ConvLSTM \citep{kondylatos2022wildfire}, GRU, Conv-GRU, U-TAE, FireCastNet \citep{michail2025firecastnet}, CNN+ConvLSTM, U-Net+ConvLSTM \citep{huot2020deep} \\
\midrule
& & & & Pixel-wise
& XGBoost \citep{pelletier2023wildfire}\\
\multirow{-2}{*}{\makecell[l]{Satellite Bands}}
& \multirow{-2}{*}{\makecell[l]{National}}
& \multirow{-2}{*}{\makecell[l]{50 m -- 8 km}}
& \multirow{-2}{*}{\makecell[l]{Weekly}}
& Spatio-Temporal
& CNN+LSTM \citep{yang2021predicting}\\
\midrule
\rowcolor[HTML]{A4D9A2}
\makecell[l]{Multi-Modal:\\ CanadaFireSat}
& \makecell[l]{Canada}
& \makecell[l]{100 m}
& \makecell[l]{8 days}
& Spatio-Temporal
& ResNet+ConvLSTM, ViT+ConvLSTM \\
\bottomrule
\end{tabular}
}
\caption{Comparison of some existing wildfire prediction methods across predictor types with CanadaFireSat.}
\label{tab:comp}
\end{table}

The growing availability of open-access remote sensing data \citep{reichstein2019deep, camps2021deep}, which enables the monitoring of large and remote regions, now allows mapping the drivers of fire susceptibility across multiple sensors like MODIS and weather data \citep{ghaderpour2025analyses} or combined MODIS, weather data, and Sentinel-2 \citep{dadkhah2025analyzing}. This accumulation of data contributed to the emergence of wildfire forecasting models leveraging remote sensing imagery with neural networks \citep{xu2025deep}. Table \ref{tab:comp} provides an overall view of wildfire forecasting methods across predictors and method types. When processing hydrometeorological data in the form of one-dimensional inputs (i.e. tabular data), the method of choice is the Multi-Layer Perceptron (MLP): \citet{buch2023smlfire1} models fire frequencies and sizes across the western US, while \citet{joshi2021improving} predicts global burned area. For temporal series of tabular inputs, Long Short Term Memory (LSTM) networks have been proposed due to their ability to capture temporal relationships, with applications ranging from fire prediction in India \citep{natekar2021forest} to daily fire danger forecasting at regional scales \citep{prapas2021deep, kondylatos2022wildfire}. When considering spatial inputs, convolutional neural networks (CNN) \citep{prapas2022deep} and vision transformer (ViT) \citep{prapas2023televit} incorporating teleconnections for sub-seasonal forecasting, have been explored. Finally, for spatio-temporal data, architectures like convolutional LSTM (ConvLSTM), which join the sequence processing abilities of LSTM networks to the spatial awareness of CNNs, have been applied to next-day fire danger forecasting in the Mediterranean \citep{prapas2021deep, kondylatos2022wildfire}, fire spread prediction across the U.S. \citep{huot2020deep}, and peatland fire prediction in Canada \citep{bali2021prediction}. There is no clear consensus on the best model to use, as results seem to vary depending on the dataset characteristics, region of interest, forecast horizon, and predictors list.\citep{jain2020review} highlight this in their broad review, while \citep{huot2020deep} and \citep{prapas2021deep} reach different conclusions about the relative importance of spatial versus temporal context when comparing similar architectures over the US and Greece, respectively. In terms of data, most methods leverage hydrometeorological predictors, from reanalysis data like ERA5 to weather stations, with spatial resolutions ranging from $\sim 27$ km to $4$ km. Finally, researchers resort to remote sensing products (characterizing the vegetation) at higher resolution (up to $500$ m) and static factors symptomatic of land cover type and human activities. The individual predictors are then re-sampled to the target resolution corresponding to the final wildfire probability map, varying from $0.25$° for global applications \citep{bakke2023data, prapas2022deep} to up to $1$ km for localized regions \citep{kondylatos2022wildfire, huot2020deep, prapas2021deep}, with higher resolution approaches ($50$ m, $30$ m) either point-based \citep{pelletier2023wildfire} or extremely local with no temporal dimension \citep{jaafari2019hybrid}. For instance, in the context of large countries such as Canada, which spans thousands of kilometers, the output resolution of current wildfire probability maps is $\sim 0.1$° \citep{bali2021prediction}. This represents an important limitation, as such coarse wildfire probability maps prevent wildfire management from properly allocating resources at a finer scale and lead to the underestimation of potential smaller wildfires.

\begin{figure}[htpb]
    \centering
    \includegraphics[width=\textwidth]{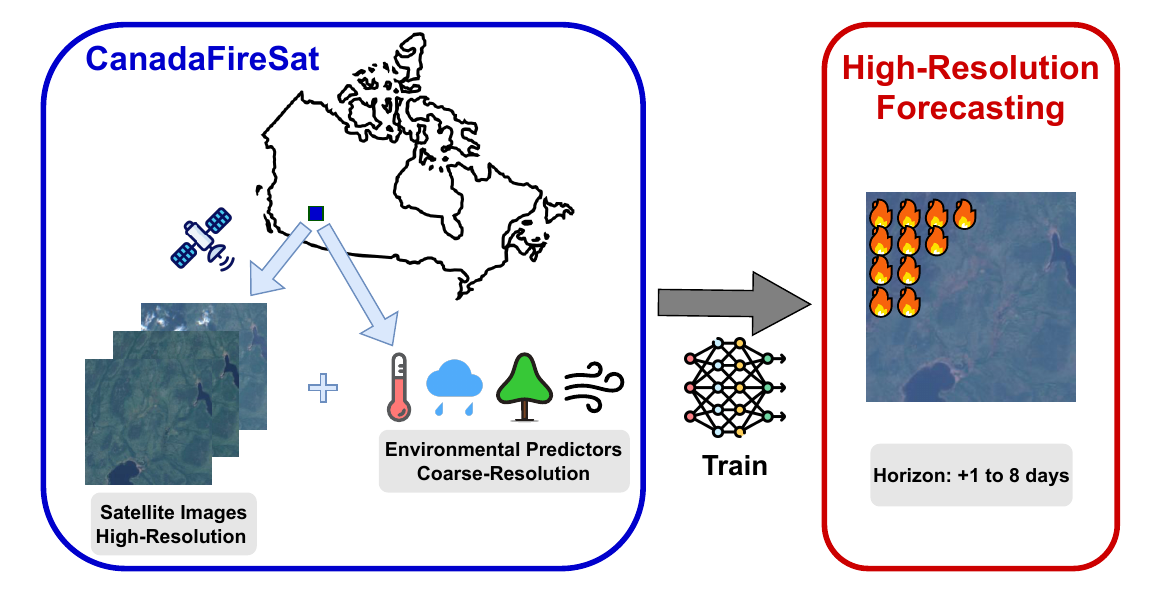}
    \caption{The CanadaFireSat benchmark and the high-resolution wildfire forecasting task.}
    \label{fig:sum}
\end{figure}

In this paper, we propose a novel multi-modal and spatio-temporal dataset covering Canada. CanadaFireSat enables high-resolution wildfire forecasting, addressing the lack of standardized benchmarks identified in the literature \citep{ejaz2025comprehensive}, and provides a unified framework for evaluating and comparing different predictive models. High-resolution wildfire forecasting is defined in Section \ref{sec:methods} as a binary patch classification task, with  $100$ m resolution patches:  a patch is labeled with the class \textit{fire} if it contains at least one burned pixel during an 8-day window at the native label resolution of Sentinel-2 ($10$ m). Our contributions are as follows:

\begin{enumerate}
    \item We introduce a benchmark dataset, CanadaFireSat, available on the \href{https://huggingface.co/datasets/EPFL-ECEO/CanadaFireSat}{Hugging Face Hub}, for high-resolution wildfire forecasting at $100$ m over Canada in 8-day forecasting window. CanadaFireSat enables high-resolution wildfire forecasting by resorting to temporal series of multi-spectral images (Sentinel-2) complemented by temporal series of environmental drivers from both reanalysis data (ERA5) and coarse resolution satellite products (MODIS), as shown in Figure \ref{fig:sum}.
    \item We investigate the impact of negative sampling on wildfire forecasting through the collection of two test sets across the 2023 extreme wildfire season for CanadaFireSat. Besides a classic test set following the same sampling strategy as the train and validation sets, where wildfire forecasting models show compelling performance, we also propose a hard test set sampled adversarially: this allows studying the lower-bound performance of models under extreme conditions, where ignition constitutes the key discriminating factor to identify potential wildfires.
    \item We demonstrate the potential of learning multi-modal models for high-resolution wildfire forecasting by benchmarking two state-of-the-art computer vision architectures on CanadaFireSat: ResNet \citep{he2016deep} and ViT \citep{dosovitskiy2020image} across three settings with varying input modalities: \One (Sentinel-2 at $10$ m), \Two (ERA5 at $11$ km, FWI at $0.25$°, MODIS at $1$ km and $500$ m), and \Three.
\end{enumerate}

CanadaFireSat allows a big leap in terms of resolution with respect to what was possible with previous datasets, such as \citep{huot2020deep} or \citep{prapas2021deep}, both using a target resolution of $ 1$ km over the U.S. and the Eastern Mediterranean region, respectively. 
Moreover, our results on CanadaFireSat demonstrate that: \textbf{i)} deep learning models outperform a knowledge-driven baseline (FWI) in both normal and extreme fire seasons, and \textbf{ii)} multi-spectral and hydrometeorological data complement each other, with multi-modal models providing the most accurate predictions, even outperforming uni-modal deep learning-based models \citep{prapas2023televit, michail2025firecastnet, yang2021predicting}.

\begin{table}[h]
    \centering
    \begin{tabular}{|l|c|}
        \hline
        \textbf{Statistic} & \textbf{Value} \\
        \hline
        Total Samples & 177,801 \\
        Target Spatial Resolution & 100 m \\
        Region Coverage & Canada \\
        Temporal Coverage & 2016 - 2023 \\
        Sample Area Size & $2.64 \; \text{km} \times 2.64 \; \text{km}$  \\
        Fire Occurrence Rate & 39\% of samples\\
        Total Fire Patches & 16\% of patches \\
        Training Set (2016-2021) & 78,030 samples \\
        Validation Set (2022) & 14,329 samples \\
        Test Set (2023) & 85,442 samples \\
        Sentinel-2 Temporal Median Coverage & 55 days (8 images) \\
        Number of Environmental Predictors & 58 \\
        Data Sources & ERA5, MODIS, CEMS \\
        \hline
    \end{tabular}
    \caption{Main Statistics of the CanadaFireSat Dataset}
    \label{tab:CanadaFireSat_stats}
\end{table}

\section{The CanadaFireSat Dataset}
\label{sec:dataset}

In this section, we present CanadaFireSat, a benchmark dataset for high-resolution wildfire forecasting. First, we describe the sampling scheme for the selection of positive and negative data samples in Section \ref{sec:sample}. Then, in Section \ref{sec:pred} we detail the set of predictors extracted and combined to build our multi-modal learning benchmark for high-resolution wildfire forecasting. Table \ref{tab:CanadaFireSat_stats} summarizes CanadaFireSat's main characteristics.

\subsection{Sample Identification}
\label{sec:sample}

Covering the entirety of Canada with Sentinel-2 images at $10$ m requires extremely high storage capacity, beyond the size of typical datasets. As such, to represent all territories and provinces of Canada, we build CanadaFireSat by resorting to a sampling strategy. As our fire labels are binary, we sample the dataset as a series of positive and negative examples. For fire (\textit{positive}) sample identification, we first extract all fires that occurred between 2015 and 2023, as described in Section \ref{sec:bu}. Then, samples not including any fire event (\textit{negative}) are sampled across the same period for all provinces and territories, depending on their FWI and acquisition dates, as detailed in Section \ref{sec:unbu}.

\begin{figure}[htpb]
    \centering
    \includegraphics[width=0.65\textwidth]{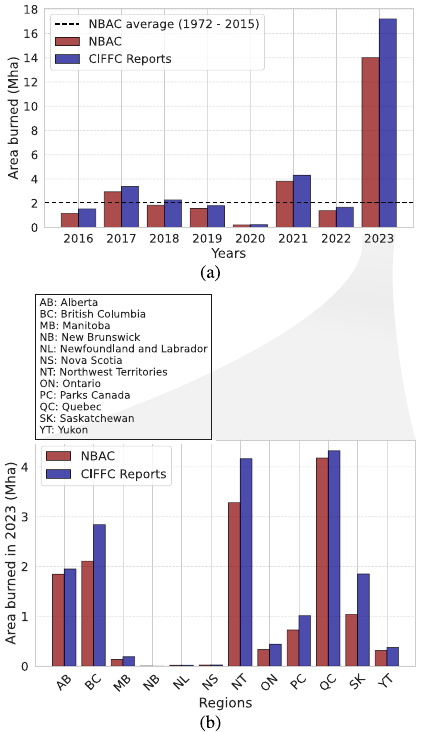}
    \caption{Burned area in Canada (millions of hectares) from NBAC compared to Canadian Interagency Forest Fire Centre (CIFFC) data. \textbf{(a)} Annual burned area from 2016 to 2023. Discrepancies between sources stem from unstandardized CIFFC reporting across territorial agencies and differing timelines: CIFFC provides near real-time estimates while NBAC is compiled up to 6 months post-year. \textbf{(b)} Per-region burned area for 2023, with Québec, Northwest Territories (which includes Nunavut in NBAC statistics), and British Columbia most impacted, and also the regions with the largest inter-source discrepancies.}
    \label{fig:burn}
\end{figure}

\subsubsection{Positive Samples}
\label{sec:bu}

Fire samples in our CanadaFireSat dataset are identified based on the fire polygons of the National Burned Area Composite\footnote{Available at \url{https://cwfis.cfs.nrcan.gc.ca/datamart/metadata/nbac}} (NBAC) \citep{hall2020generating} from the Canadian National Fire Database. NBAC has been compiled annually since 1972 and integrates data from Natural Resources Canada, provincial and territorial agencies, and Parks Canada, using a rule-based approach to select the most accurate data source to delineate the burned area perimeters; this includes ground and aerial surveys or post-event satellite imagery analysis from Landsat (5, 6, 7, 8, 9 or MSS), Sentinel-2, MODIS, VIIRS, and AVHRR. We focus on all fires since 2015 (the launch of the first Sentinel-2 satellite) up to 2023, with no restriction on ignition sources or other fire metadata. Over this time, a large majority of the polygons were compiled from ground survey, Landsat, aerial survey, and Sentinel-2 in this respective order. In Figure \ref{fig:burn}a, we report the NBAC yearly average burned area for this period, with 2022 reaching $1.38$ mha burned and 2023 reaching $14.01$ mha burned. This outlines the difference in wildfire season severity for our validation (2022) and test (2023) sets compared to the average from 1972 to 2015 of $\sim 2.03$ mha. In other words, 2023 was an exceptional fire season.

\begin{figure}[htbp]
 \captionsetup[subfigure]{justification=centering}
  
  \begin{subfigure}{\textwidth}
    \centering
    \includegraphics[width=0.6\textwidth]{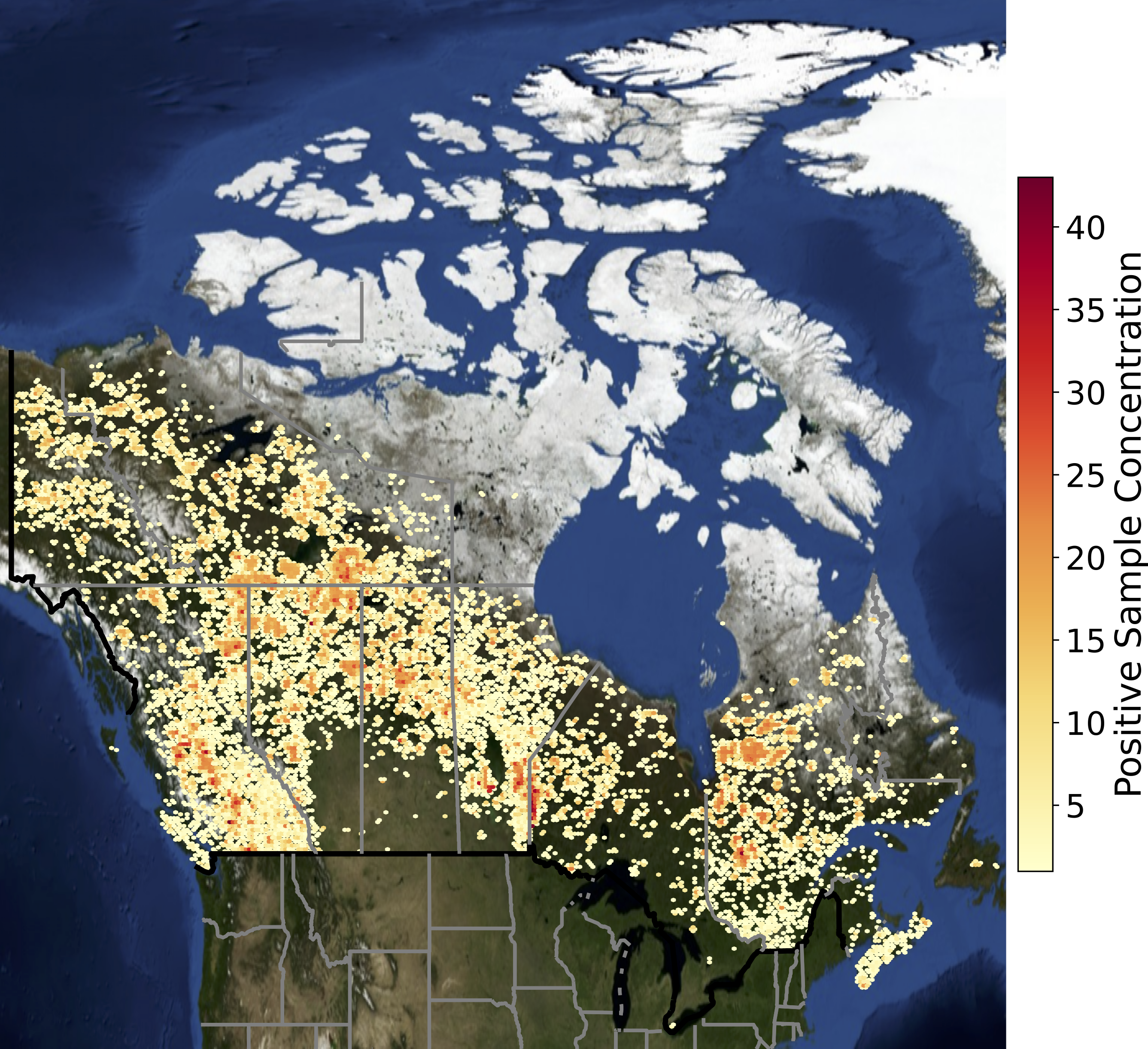}
    \subcaption{Positive sample distribution across the period 2015-2023}
    \label{fig:pos-sample}
  \end{subfigure}
  
  \vspace{0.5cm} 
  
  \begin{subfigure}{\textwidth}
    \centering
    \includegraphics[width=0.6\textwidth]{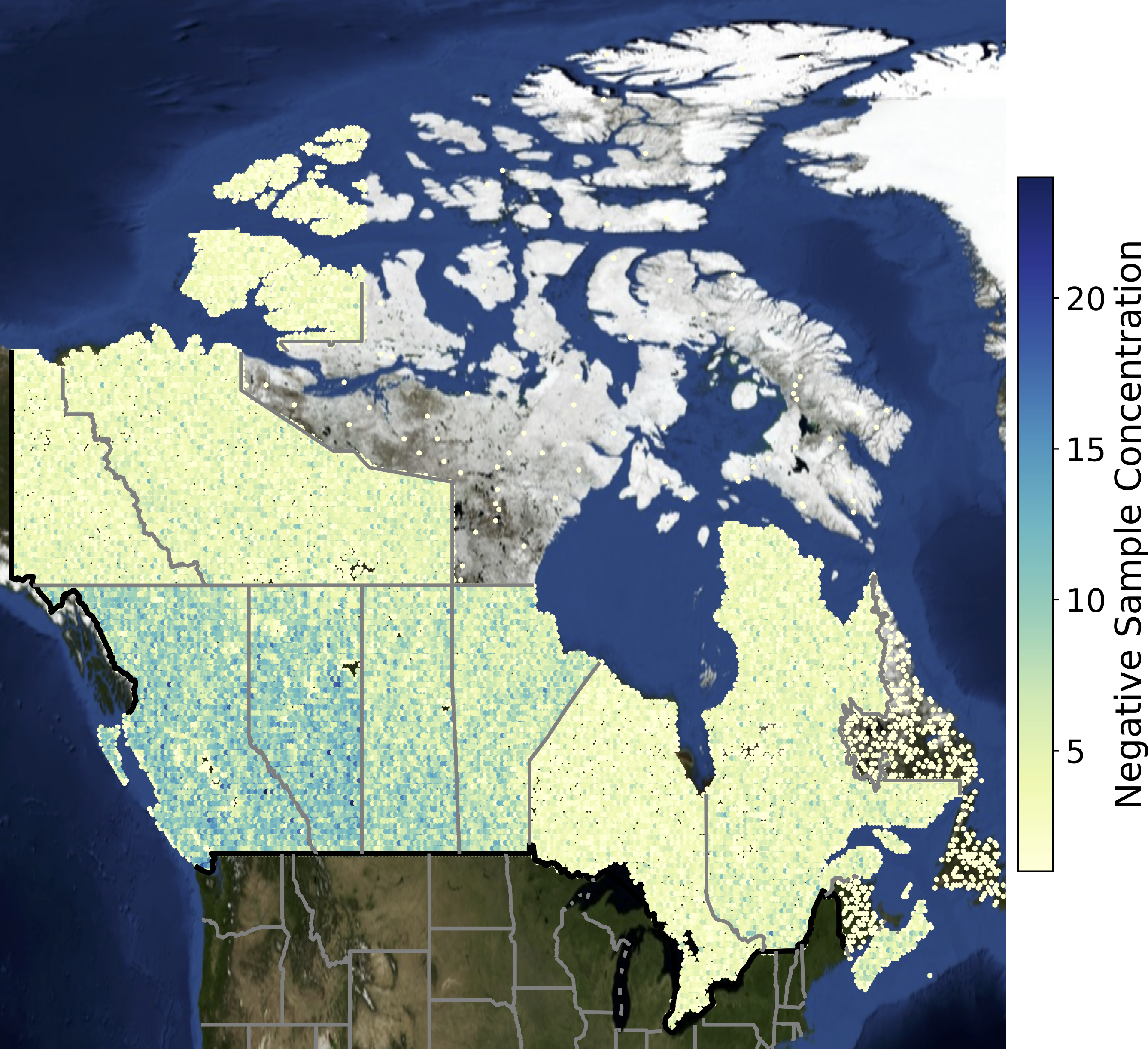}
    \subcaption{Negative sample distribution across the period 2015-2023}
    \label{fig:neg-sample}
  \end{subfigure}

  \caption{Distribution of positive (containing burned area) and negative samples (following our FWI-based sampling strategy) from 2015-2023, before any post-processing.}
  \label{fig:sample}
\end{figure}

Positive samples for the CanadaFireSat dataset are extracted from the NBAC fire polygons through two aggregation processes. First, through a spatial aggregation on a $2.8 \; \text{km} \times 2.8 \; \text{km}$ grid over Canada, where positive samples are identified as the grid entries intersecting the fire polygons. We used a small buffer around the $2.64 \; \text{km} \times 2.64 \; \text{km}$ Sentinel-2 tiles to avoid any potential overlap between samples due to imprecision in the data processing during large wildfires (Figure \ref{fig:use case}) and ensuring sample independence. The chosen image size follows standard practices in Earth observation benchmarks, considering Sentinel-2 data \citep{manas2021seasonal, wang2022ssl4eo}. While other benchmarks consider smaller tiles~\mbox{\citep{sumbul2019bigearthnet,helber2019eurosat}}, our spatial extent is chosen to balance contextual information for wildfire forecasting against computational constraints imposed by the addition of the temporal dimension in the data cubes. Second, a temporal aggregation is performed in two steps: 1) all fires temporally overlapping inside a grid entry are accounted as a single fire occurring from the first fire start date to the last fire end date, and 2) leveraging the 8-day temporal grid from products such as NDVI from MODIS (starting each year at the 1st of January) we aggregate all fires within a spatial grid entry occurring during the same 8-day window. This is done to build our 8-day wildfire forecasting benchmark where, for a given time-step $t$, our model should predict the probability of a fire occurring in the next 8 days, i.e. from $t$ to $t + 7$ included, leveraging predictors (both satellite and environmental, see Section \ref{sec:pred}) from $t - \Delta t$ to $t - 1$. In Figure \ref{fig:pos-sample}, we showcase the spatial distribution of positive samples across Canada, for a total of $n_{\text{pos}} = 88,110$ samples before any post-processing (detailed in Section \ref{sec:pred}). Outside of British Columbia, most fires occur in the boreal ecosystem. This pattern is very visible across Alberta, Saskatchewan, and Manitoba, where the Great Plains in the southern portions of these provinces show little to no fires compared to the boreal forest in the north.

\subsubsection{Negative Samples}
\label{sec:unbu}

As we aim to build our benchmark on multi-modal inputs, including satellite image time series, we are limited in disk storage to densely sample Canada over the whole period from 2015 to 2023. Therefore, we sample a negative set of size $n_{\text{neg}} = 2 \cdot n_{\text{pos}}$ to match the degree of imbalance of other wildfire forecasting datasets \citep{huot2020deep, prapas2021deep, kondylatos2022wildfire}. We sample from the same grid defined in Section \ref{sec:bu}, $G_{y, r}$, for each year $y$ between the first and last fires during that year (so beyond the wildfire season), and across all regions $r$. For a given year $y > 2015$ and region $r$, we avoid locations where a fire occurred in the previous years: $ \bigcup_{i=2015}^{y-1} F_{i,r}$, or locations that were already selected as negative samples in the previous years: $\bigcup_{i=2015}^{y-1} N_{i,r}$. Our negative set for a given region and year can be defined as:

\begin{equation}
    N_{y,r} \sim S_{y, r} = \{x \in G_{y,r} | \; x \notin \bigcup_{i=2015}^{y-1} F_{i,r} \wedge x \notin \bigcup_{i=2015}^{y-1} N_{i,r}\} \: , 
    \label{eq:sample}
\end{equation}

where $N_{y, r}$ is the set of negative samples and $S_{y,r}$ the set of potential locations in the grid. We sample $N_{y, r}$ uniformly across levels (defined by decile bins) of the FWI:

\begin{equation}
    P_{\text{FWI}}(x | N_{y,r}) \propto P_{\text{FWI}}(x | S_{y,r}) \: .
\end{equation}

In practice, this is done by partitioning the FWI distribution into ten decile bins: $[B_1, \dots, B_{10}]$ across the FWI quantiles $[Q_1, \dots, Q_9]$ such that each bin contains approximately $10\%$ of the observations, and uniformly sampling across those decile bins for $N_{y, r}$. Each bin $B_l$ is defined as a subset of the FWI range:

\begin{equation}
    B_l = \left\{ x \in \text{FWI} \,\middle|\, Q_{l-1} < x \le Q_l \right\}, \quad \text{for} \: l = 1, \dots, 10 \: ,
\end{equation}

where $Q_0 = 0$, and $Q_{10} = + \text{inf}$ are the bounds of the FWI range. This way, the negative population is representative of all fire weather conditions for each region and year, including cases where a high FWI was predicted, but no fire was observed.

Figure \ref{fig:neg-sample} presents the spatial distribution of the sampled negative locations across all years: it shows that, per region, the negative samples are well spread spatially, contrary to the positive samples, as we aim to represent the complete patterns of fire danger conditions. British Columbia, Alberta, Saskatchewan, and Manitoba contain the highest concentration of negative samples in certain areas due to the high concentration of fires in those regions (negative samples are sampled twice as much as positive ones). On the contrary, Nunavut, Newfoundland and Labrador, and New Brunswick are less densely sampled due to a lack of fires during the analyzed period. We select in total $n_{\text{neg}} = 176,650$ negative samples that, combined with our positive samples $n_{\text{pos}}$, consitute the CanadaFireSat Train (2016 - 2022), Val (2022), and Test (2023) sets. Note that some of these samples will be filtered out through the post-processing procedure described in Section \ref{sec:pred}.

\begin{figure}[!t]
  \captionsetup[subfigure]{justification=centering}

  \centering

  \begin{subfigure}{0.48\textwidth}
    \centering
    \includegraphics[width=\linewidth]{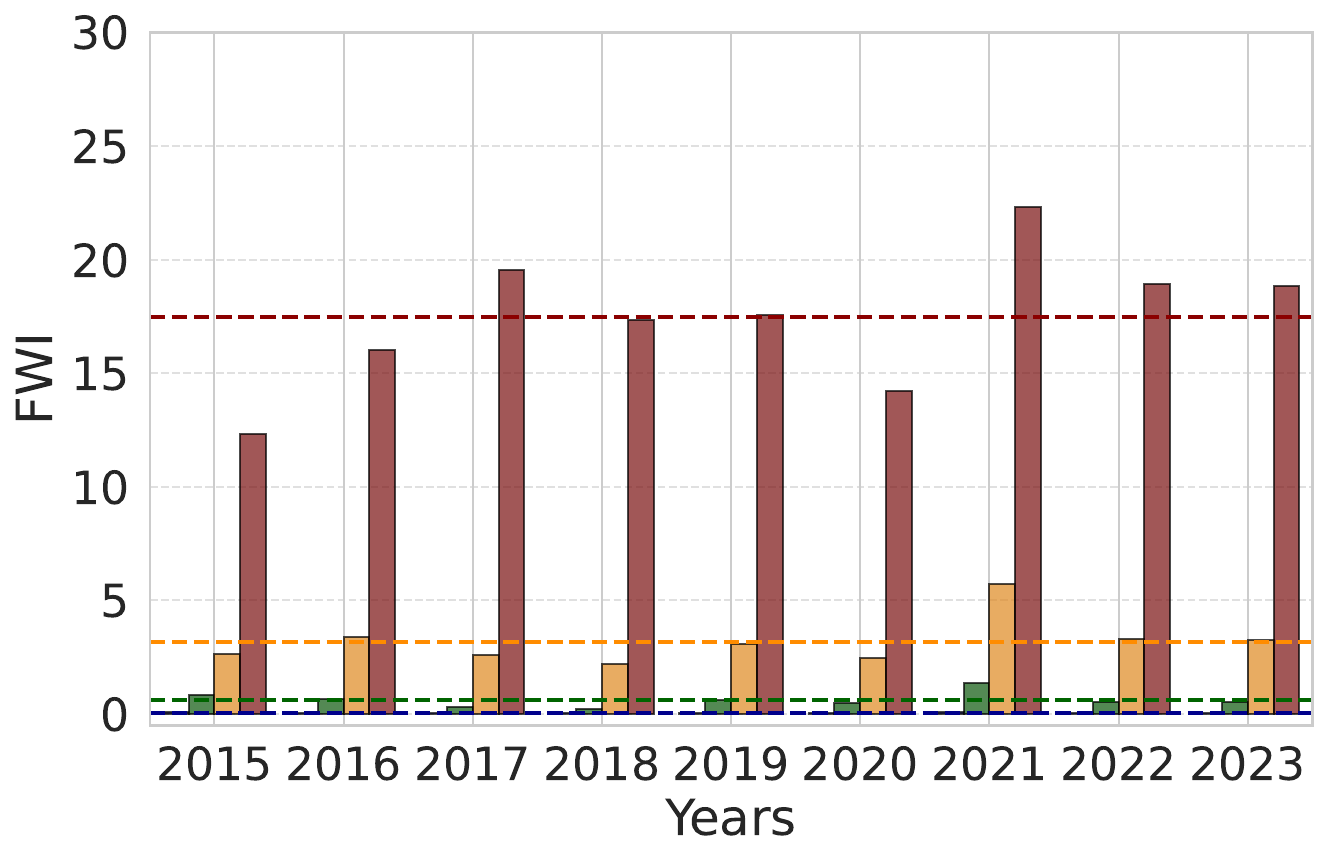}
    \subcaption{Canada}
    \label{fig:year-fwi}
  \end{subfigure}
  \begin{subfigure}{0.48\textwidth}
    \centering
    \includegraphics[width=\linewidth]{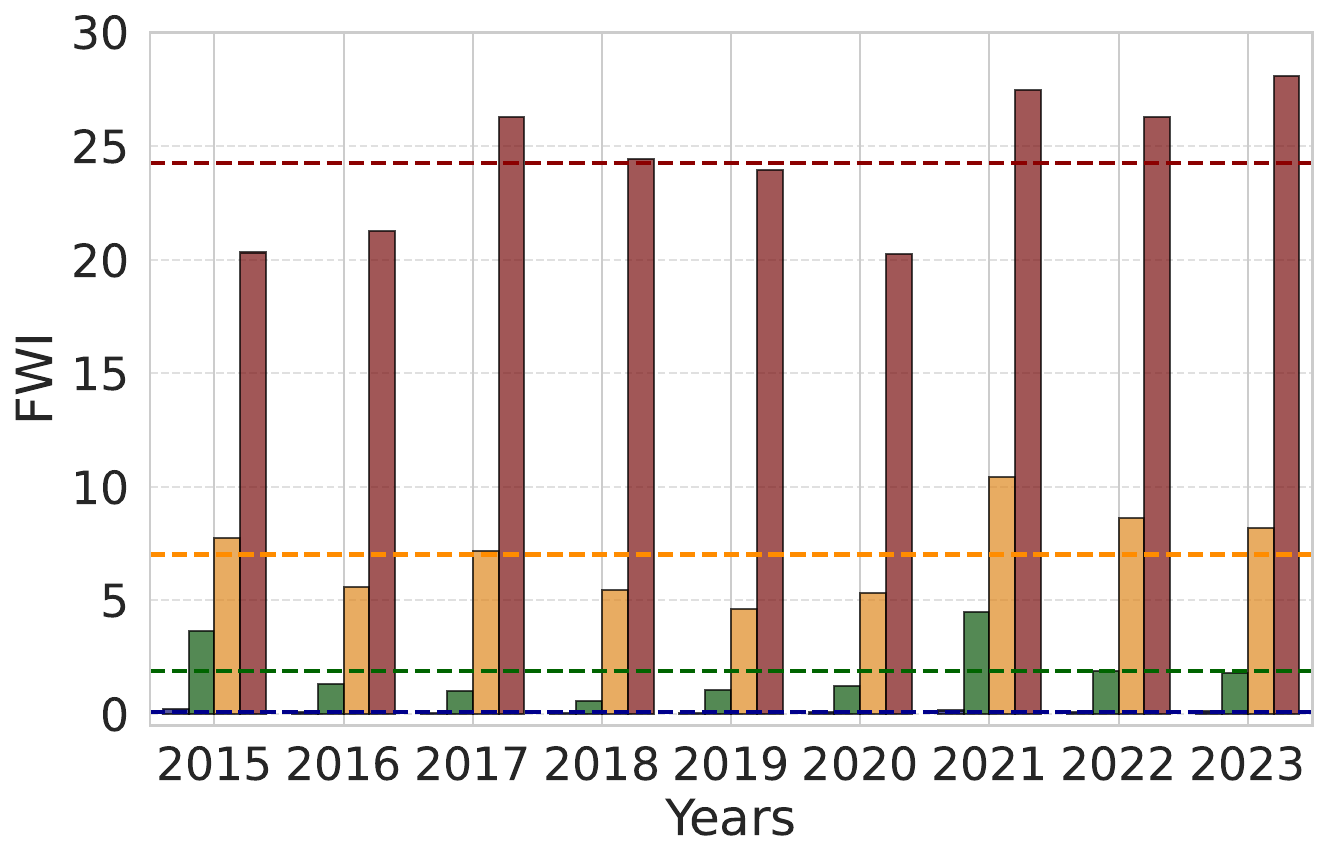}
    \subcaption{Alberta}
    \label{fig:alb-fwi}
  \end{subfigure}

  \begin{subfigure}{0.5\textwidth}
    \centering
    \includegraphics[width=\linewidth]{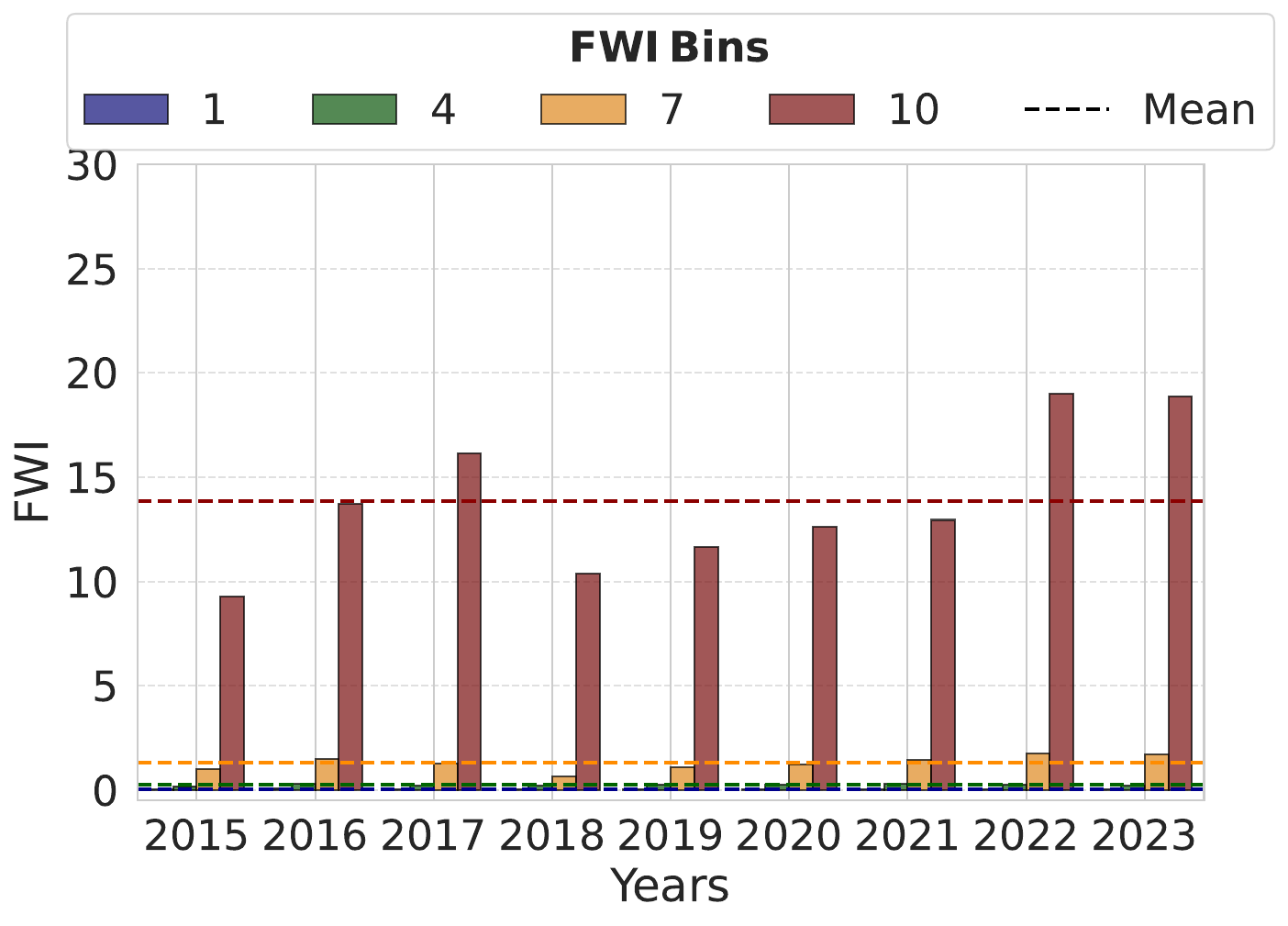}
    \subcaption{Northwest Territories}
    \label{fig:nt-fwi}
  \end{subfigure}

  \caption{Annual FWI mean over four decile bins: $\{1, 4, 7, 10\}$ across Canada, Alberta, and Northwest Territories.}
  \label{fig:fwi}
\end{figure}

In Figure \ref{fig:year-fwi}, we present the annual FWI mean for the negative sample set. We see that up to the decile bin number $4$ with a FWI mean: $\overline{\text{FWI}} = 0.62$, most negative samples will have an FWI close to $0$, as the FWI distribution of available locations for negative samples consistently presents an important peak in this range. This is representative of the FWI conditions across all regions of Canada between the first and last fires of each year. 
Furthermore, we show in Figure \ref{fig:alb-fwi} and \ref{fig:nt-fwi} that across two commonly impacted regions by wildfires (Alberta and the Northwest Territories), there are strong differences in FWI decile bin mean value between regions, with the delta for the top decile bin reaching up to $\sim 14$ in 2021. This can be explained by the higher latitude of the Northwest Territories compared to Alberta and the presence of permafrost in their northernmost areas.

We also observe a strong inter-annual variability between  2022 and 2023, as the latter was a record-breaking wildfire season in Canada \citep{jain2024drivers}, resulting in $19.6 \%$ of fire patches in the Test set compared to $11.5 \%$ in the Val set. This distribution shift shows that, despite similar fire weather conditions as presented in Figure \ref{fig:year-fwi}, wildfire occurrence is significantly higher in the Test set compared to the Training set. This can lead to the overestimation of the performance of wildfire forecasting models: by looking at the distribution of positive and negative samples in Figure \ref{fig:fwi-dist-comp}, one can observe that the FWI alone is a highly discriminative feature for the class \textit{fire} (see Section \ref{sec:perf}). As a result, we introduce an adversarial sampling strategy for the negative samples to study the lower-bound performance of wildfire forecasting models for the extreme year $2023$, named \emph{Test Hard}. In this adversarial test set, we aim to make the distribution of the negative population similar to that of the positive population with respect to the FWI, making ignition the main discriminative factor. To sample negative samples for Test Hard, we perform a stratified sampling for the year 2023 in the following way. First, by extending Equation \ref{eq:sample} to account for both land cover and the month of the year. Then, for a given land cover $c$ and month of the year $m$, we sample $N_{y, r, m, c}$ uniformly across levels (defined by decile bins) of the FWI for the positive samples population $F_{y, r, m, c}$ :

\begin{figure}[!t]
 \captionsetup[subfigure]{justification=centering}
  \begin{subfigure}{\textwidth}
    \centering
    \includegraphics[width=0.8\textwidth]{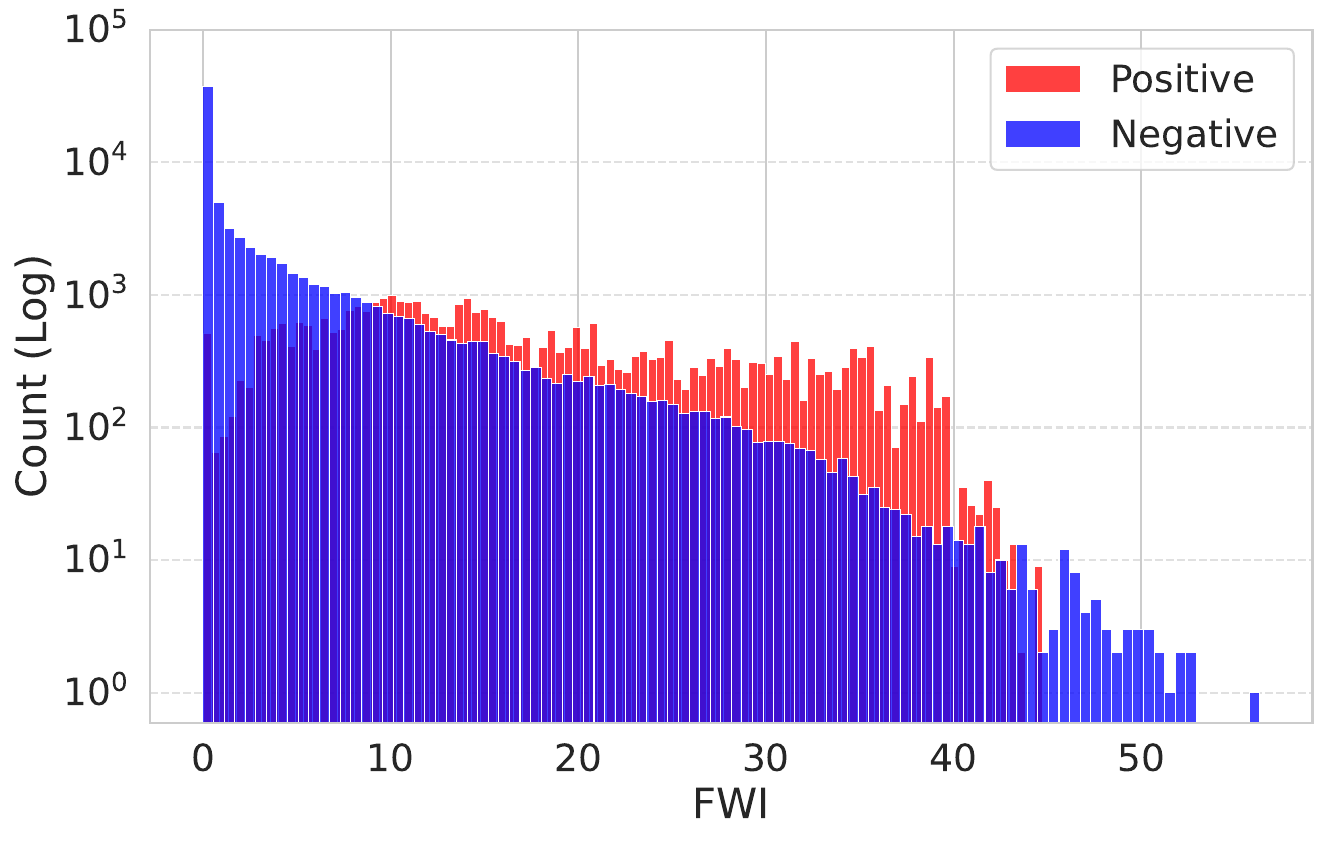}
  \subcaption{FWI distribution in log-scale for the Test set across positive and negative samples.}
  \label{fig:fwi-dist}
  \end{subfigure}
  
  \vspace{0.5cm} 

  \begin{subfigure}{\textwidth}
    \centering
    \includegraphics[width=0.8\textwidth]{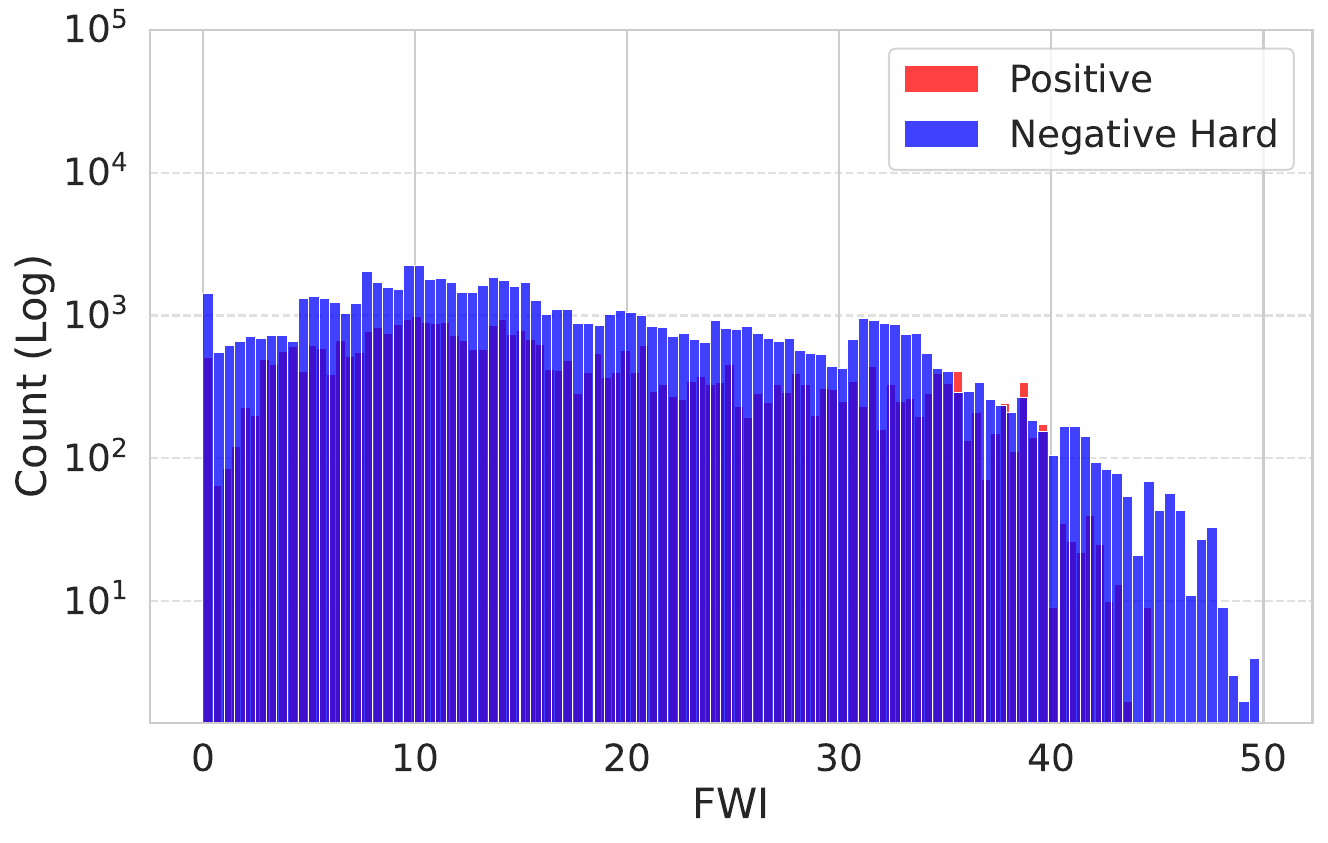}
  \subcaption{FWI distribution in log-scale for the Test Hard set across positive and negative samples.}
  \label{fig:fwi-dist-hard}
  \end{subfigure}
  \caption{Comparison of the FWI distribution in log-scale across the Test and Test Hard sets for both positive and negative samples.}
\label{fig:fwi-dist-comp}
\end{figure}

\begin{equation}
    P_{\text{FWI}}(x | N_{y, r, m, c}) \propto P_{\text{FWI}}(x | F_{y, r, m, c}) \: ,
\end{equation}

and sample $n_{\text{neg}}(y, r, m, c) \simeq 2 \times n_{\text{pos}}(y, r, m, c)$ negatives. The land cover is downloaded from ESA WorldCover at $10$ m for 2020. The resulting distribution is shown in Figure \ref{fig:fwi-dist-comp} and represents 77,247 complementary negative samples to CanadaFireSat statistics reported in Table \ref{tab:CanadaFireSat_stats}. By deploying the trained networks on Test Hard, where ignition acts as the main discriminative factor, we can assess their performance on modeling this complex triggering factor whose patterns can only be implicitly learned from the training data. For this reason, the performance of our trained models for high-resolution wildfire forecasting on Test Hard can be considered as a lower bound for such an extreme wildfire season as presented in Section \ref{sec:perf}. Further details about the distribution of samples across land-cover classes are provided in Figure \ref{fig:lc-dist-comp}.

\subsection{Predictors}
\label{sec:pred}

The predictors used in CanadaFireSat fall into two categories: satellite image time series and environmental data. 

\subsubsection{Satellite Image Time Series}
\label{sec:sat}

\begin{figure}[htbp]
\centering
\setlength\tabcolsep{1.5pt} 
\resizebox{\columnwidth}{!}{
 \begin{tabular}{lcccc}
 
 \rotatebox{90}{\parbox{3cm}{\centering S-2 Input}} & 
 \includegraphics[width=0.24\textwidth, height=3cm]{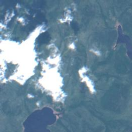} & 
 \includegraphics[width=0.24\textwidth, height=3cm]{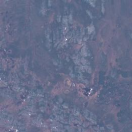} & 
 \includegraphics[width=0.24\textwidth, height=3cm]{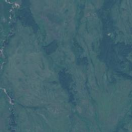} &  
 \includegraphics[width=0.24\textwidth, height=3cm]{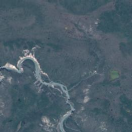} \\

 \rotatebox{90}{\parbox{3cm}{\centering S-2 Input}} & 
 \includegraphics[width=0.24\textwidth, height=3cm]{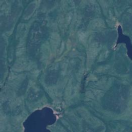} & 
 \includegraphics[width=0.24\textwidth, height=3cm]{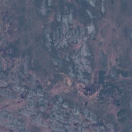} & 
 \includegraphics[width=0.24\textwidth, height=3cm]{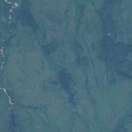} &  
 \includegraphics[width=0.24\textwidth, height=3cm]{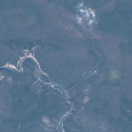} \\

 \rotatebox{90}{\parbox{3cm}{\centering S-2 Input}} & 
 \includegraphics[width=0.24\textwidth, height=3cm]{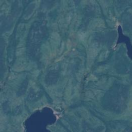} & 
 \includegraphics[width=0.24\textwidth, height=3cm]{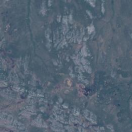} & 
 \includegraphics[width=0.24\textwidth, height=3cm]{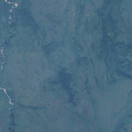} &  
 \includegraphics[width=0.24\textwidth, height=3cm]{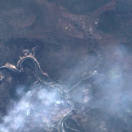} \\

 \rotatebox{90}{\parbox{3cm}{\centering S-2 Post-Fire}} & 
 \includegraphics[width=0.24\textwidth, height=3cm]{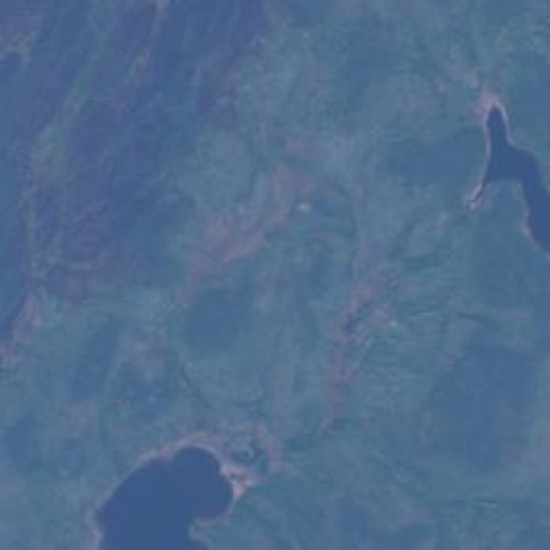} & 
 \includegraphics[width=0.24\textwidth, height=3cm]{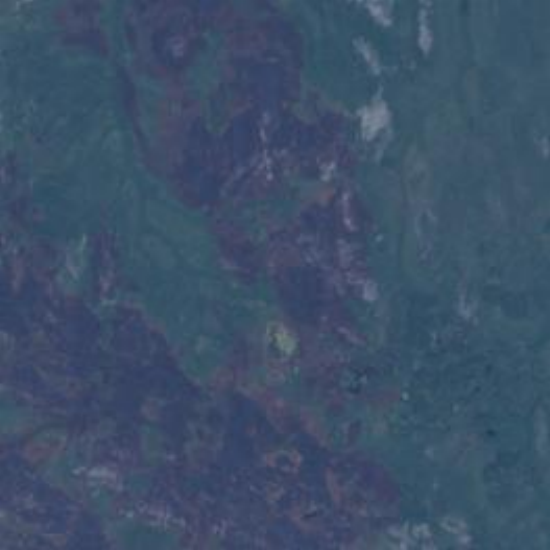} & 
 \includegraphics[width=0.24\textwidth, height=3cm]{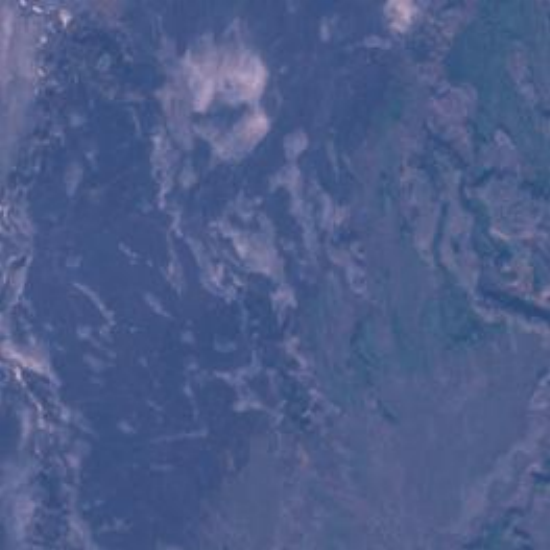} &  
 \includegraphics[width=0.24\textwidth, height=3cm]{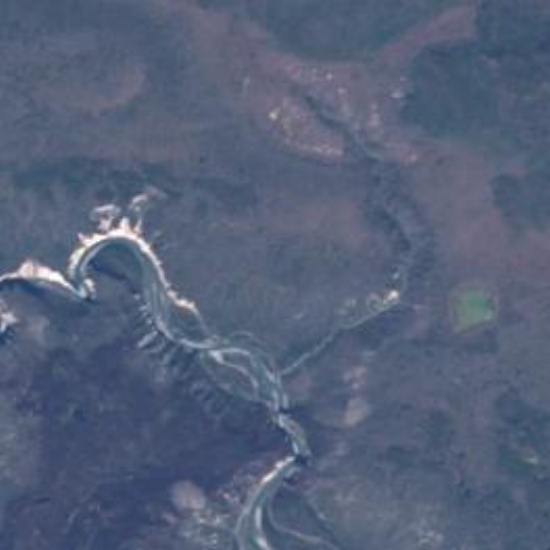} \\

 \rotatebox{90}{\parbox{3cm}{\centering S-2 and Label}} & 
 \includegraphics[width=0.24\textwidth, height=3cm]{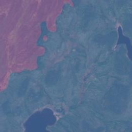} & 
 \includegraphics[width=0.24\textwidth, height=3cm]{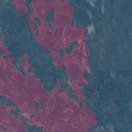} & 
 \includegraphics[width=0.24\textwidth, height=3cm]{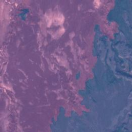} &  
 \includegraphics[width=0.24\textwidth, height=3cm]{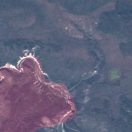} \\

 \end{tabular}}
 \caption{\textbf{Row 1-3} Samples of Sentinel-2 input time series for four locations in Canada. We show only the RGB bands at $10$ m resolution with rescaled intensity. \textbf{Row 4} Sentinel-2 images after the fire occurred. \textbf{Row 5} Fire polygons used as labels with the Sentinel-2 images post-fire as background.}
 \label{fig:sat}
\end{figure}

To be able to forecast wildfires at a patch resolution of $100$ m, we need high-resolution information. However, hydrometeorological fire danger predictors cannot be found at $100$ m resolution for the entirety of Canada. Therefore, we investigate the potential of multi-spectral high-resolution satellite images as proxies for fire predictors following previous literature \citep{pelletier2023wildfire, yang2021predicting}. We use the 13 bands from Sentinel-2 (S2) L1C harmonized data as proxies to several known fire predictors, such as NDVI or soil moisture. We use the L1C products as they are directly available for the whole period 2015-2023 without any need for further processing. Moreover, we extract temporal data to better estimate the impact of changes in the hydrometeorological conditions on the local ecosystem. Top-of-atmosphere reflectance from Sentinel-2 is impacted by aerosols, clouds, topography effects, and other phenomena that can bias its measurement across the multi-spectral bands for numerous land cover types \citep{sola2018assessment}, in particular for shorter wavelengths like the RGB bands. This can impact the computation of radiometric indices often used in burned area mapping \citep{howe2022comparing} and the precision of non-local and multi-temporal analyses of Sentinel-2 data. However, we decided to still rely on L1C products for the following reasons. First, since the beginning of the Sentinel-2 mission in 2015 up until 2017, global data were only available at the level Top of Atmosphere: L1C\footnote{\url{https://docs.sentinel-hub.com/api/latest/data/sentinel-2-l1c/}}. L2A processing was initially limited to some regions (release of L2A data started in January 2016), and Surface Reflectance data are globally available since January 2017\footnote{\url{https://docs.sentinel-hub.com/api/latest/data/sentinel-2-l2a/}}. Moreover, in Google Earth Engine, which we use as our harmonization data platform, L2A products are only available from 2019\footnote{\url{https://developers.google.com/earth-engine/datasets/catalog/COPERNICUS_S2_SR_HARMONIZED}}. Therefore, using L2A products would have limited the temporal coverage of CanadaFireSat. Second, machine and deep learning can mitigate the limitations reported above by implicitly learning the approximate corrections necessary for the downstream application targeted through correction agnostic models \citep{russwurm2023large, wright2025training} or L1C specific ones \citep{medina2020machine, wright2024clouds2mask} even in the context of burned area mapping \citep{rumora2020impact}. Although specific pre-processing targeted for wildfire forecasting could reduce the data uncertainty, we consider that this is out of the scope of this work.

For a given sample $x_t \in G_{y, r}$ (the $2.8 \; \text{km} \times 2.8 \; \text{km}$ grid over Canada), we download all the full (no missing values) S2 images of size $2.64 \; \text{km} \times 2.64 \; \text{km}$ following \citep{manas2021seasonal} centered within each grid cell $x_t$ between the date $t - 64$ days and $t - 1$ day. We exclude images with a cloud cover above $40 \%$. This represents 13 images, given the average revisit time of 5 days for S2 (after the launch of Sentinel-2B). To avoid artifacts, we use a lossless compression, and we multiplied each band intensity by a factor $1e^{-4}$ to then rescale the values to 8-bit unsigned integers. Once all S2 images are extracted, a second filter on cloud coverage was applied, based on the S2 cloud probability product, but focusing only on the sample location. After this filtering, the samples $x_t$ with less than three S2 images or covering a period of less than 40 days are removed, as we aim to learn local temporal dynamics. Finally, as multiple S2 tiles can cover a sample $x_t$, we keep the tile with the most valid images for this sample. The final positive sample set after filtering is of size $69,876$ ($\sim 79\%$ of the original set), and the negative sample set is of size $107,925$ ($\sim 61\%$ of the original set). For Test Hard, the final number of negative samples is $66,406$ ($\sim 86\%$ of the previously identified samples).

The fire polygons associated with each positive sample are rasterized based on the B3 band of the S2 image that preceded the start of the fire. This process outputs binary maps of size $264 \times 264$ pixels at a resolution of $10$ m that will then be downscaled to $100$ m resolution during training. Figure \ref{fig:sat} shows examples of S2 image time series from the positive set and the expected output when a fire occurred (last row).

\subsubsection{Environmental Predictors}
\label{sec:env}

Fire weather indices and most wildfire forecast models rely on hydrometeorological drivers such as temperature, precipitation, and soil moisture. Some forecast models also leverage vegetation indices such as NDVI, EVI, or LAI. 
We include such coarse environmental predictors (summarized in Table~\ref{tab:predictors}) despite the difference in resolution between them and our target outputs, since we believe that multi-modal methods can benefit from them,  as they are strongly correlated to fire probability.


\begin{itemize}
    \item First, we extract five different MODIS products from MOD15A2H, MOD11A1, and MOD13A1 that describe the vegetation state and temperature at moderate-to-coarse-resolution: $500$ m and $1$ km. Vegetation indices are 8-day or 16-day composites, which, similarly to SeasFire \citep{prapas2022deep}, drive the temporal aggregation over 8 days of the other environmental predictors, and the NBAC burned area polygons.

    \item We also extracted 12 hydrometeorological drivers from ERA5-Land daily (detailed in Table \ref{tab:predictors}) at coarse-resolution ($11$ km), and aggregated those variables through mean, max, and min operators on the 8-day temporal grid defined by MODIS. We extend this set of predictors with three additional ones: relative humidity, vapor pressure deficit, and wind speed, computed locally from ERA5 data.

    \item Lastly, we leverage indices related to fire danger from the Copernicus Emergency Management Service (CEMS): FWI, also used in negative sampling, and drought code, both from the Canadian Forest Fire Weather Index. This data is the coarsest of all our environmental predictors with a resolution of $0.25$° for both latitude and longitude.
    
\end{itemize}

These predictors are then post-processed to set to NaN any extreme values and aligned both spatially and non-spatially with our positive and negative samples. Similar to the satellite image time series, for each sample $x_t$, we extract the environmental predictors from $t - 64$ days to $t - 1$. The non-spatial alignment is done via the weighted average of a given predictor over the target grid cell. The spatial alignment is done for each predictor by extracting a small window of data centered on $x_t$. The window size varies depending on the source resolution. We extract windows of dimension $(32, 32)$ for MODIS products at $500$ m  and $(16, 16)$ for MODIS product at $1$ km. Moreover, for ERA5-Land data, we extract windows of size $(32, 32)$, and $(13, 13)$ for CEMS. As a consequence, for a given sample $x_t$ CanadaFireSat provides spatial predictors at multiple scales covering different spatial contexts. Both alignment methods are illustrated in Figure \ref{fig:align} in \ref{sec:align}. The aim of the spatial alignment (compared to its non-spatial counterpart) is to provide a broad spatial context of fire weather conditions around the region of interest. Models trained on CanadaFireSat should consider this difference in scale across modalities, as those presented in Section \ref{sec:methods}.

\begin{table}[htbp]
    \centering
    \renewcommand{\arraystretch}{1.2}
    \makebox[\textwidth]{
    \resizebox{1.4\textwidth}{!}{ 
    \begin{tabular}{|c|c|c|c|c|c|}
        \hline
        \textbf{Dataset} & \textbf{Name} & \textbf{Units} & \textbf{Aggregation} & \textbf{Resolution} & \textbf{Source} \\
        \hline
        \multirow{5}{*}{MODIS} & NDVI & - & 16-day composite & 500 m & Google Earth Engine \\
        & EVI & - & 16-day composite & 500 m & Google Earth Engine \\
        & LST Day (1km) & K & 8-day mean, max, min & 1 km & Google Earth Engine \\
        & FPAR & - & 8-day composite & 500 m & Google Earth Engine \\
        & LAI & - & 8-day composite & 500 m & Google Earth Engine \\
        \hline
        \multirow{15}{*}{ERA5-Land} & Surface Pressure & Pa & 8-day mean, max, min & 11.1 km & Google Earth Engine \\
        & Total Precipitation Sum & m & 8-day mean, max, min & 11.1 km & Google Earth Engine \\
        & Skin Temperature & K & 8-day mean, max, min & 11.1 km & Google Earth Engine \\
        & U Component of Wind (10m) & m/s & 8-day mean, max, min & 11.1 km & Google Earth Engine \\
        & V Component of Wind (10m) & m/s & 8-day mean, max, min & 11.1 km & Google Earth Engine \\
        & Temperature (2m) & K & 8-day mean, max, min & 11.1 km & Google Earth Engine \\
        & Temperature (2m, Max) & K & 8-day mean, max, min & 11.1 km & Google Earth Engine \\
        & Temperature (2m, Min) & K & 8-day mean, max, min & 11.1 km & Google Earth Engine \\
        & Surface Net Solar Radiation Sum & J/m² & 8-day mean, max, min & 11.1 km & Google Earth Engine \\
        & Surface Solar Radiation Downwards Sum & J/m² & 8-day mean, max, min & 11.1 km & Google Earth Engine \\
        & Volumetric Soil Water Layer 1 & m³/m³ & 8-day mean, max, min & 11.1 km & Google Earth Engine \\
        & Dewpoint Temperature (2m) & K & 8-day mean, max, min & 11.1 km & Google Earth Engine \\
        & Relative Humidity & \% & 8-day mean, max, min & 11.1 km & Own Calculation \\
        & Vapor Pressure Deficit & hPa & 8-day mean, max, min & 11.1 km & Own Calculation \\
        & Wind Speed (10m) & m/s & 8-day mean, max, min & 11.1 km & Own Calculation \\
        \hline
        \multirow{2}{*}{CEMS} & Drought Code & - & 8-day mean, max, min & 0.25° ($\sim$ 28 km) & CEMS Early Warning Data Store \\
        & Fire Weather Index & - & 8-day mean, max, min & 0.25° ($\sim$ 28 km) & CEMS Early Warning Data Store \\
        \hline
    \end{tabular}
    } 
    }
    \caption{Overview of the environmental predictors.}
    \label{tab:predictors}
\end{table}

\section{Methods}
\label{sec:methods}
To demonstrate the feasibility of forecasting wildfires at $100$ m resolution, we benchmark two deep learning architectures on the proposed CanadaFireSat dataset. We chose a CNN and a Transformer as representative computer vision models, whose encodings are used to forecast wildfire probability at an 8-day horizon. To account for multi-modal interactions, models are trained in three different settings: \One (Sentinel-2), \Two (ERA5, CEMS, MODIS), and when both \Three are available. Detailed information on the settings can be found in Table \ref{tab:setting}.

\begin{table}[htbp]
\centering
\begin{tabular}{|c|c|c|c|}
\hline
\textbf{Setting} & \textbf{Source} & \textbf{Format} & \textbf{Type}\\
\hline
\multirow{1}{*}{\raisebox{-.4ex}{\includegraphics[width=2ex]{one.pdf}}} & Sentinel-2 & Spatial & Multi-Spectral Images\\
\hline
\multirow{3}{*}{\raisebox{-.4ex}{\includegraphics[width=2ex]{two.pdf}}} & MODIS & Spatial & Environmental Products\\
                       & ERA5-Land & Spatial & Climate Reanalysis\\
                       & CEMS & Spatial & Fire Indices\\
\hline
\multirow{4}{*}{\raisebox{-.4ex}{\includegraphics[width=2ex]{three.pdf}}} & Sentinel-2 & Spatial & Multi-Spectral Images\\ 
                       & MODIS & Tabular & Environmental Products\\
                       & ERA5-Land & Tabular & Climate Reanalysis\\
                       & CEMS & Tabular & Fire Indices\\
\hline
\end{tabular}
\caption{Descriptions of the modality settings for the training of the wildfire forecasting models.}
\label{tab:setting}
\end{table}

For CanadaFireSat, wildfire forecasting is framed as a binary classification task (\textit{fire} vs \textit{no fire}) at the patch level, i.e., a binary patch classification. Across our experiments, the original labels at a native resolution of $10 \; \text{m} \times 10 \; \text{m}$ are re-scaled to $100 \; \text{m} \times 100 \; \text{m}$, by labeling a patch with the binary class \textit{fire} if any pixel within the patch is labeled as burned. This design decision aims to focus on providing alerts for any size of fires at the expense of false positive pixels at the native resolution and is often used in wildfire prediction at both coarse \citep{prapas2022deep, bakke2023data} and high-resolution \citep{pelletier2023wildfire}. It is also motivated by the shortcomings of MODIS, in particular the MCD64A1 burned area product, which is recurrently used in coarse wildfire forecasting \citep{huot2020deep, rodrigues2022firo, prapas2021deep} despite underestimating burned area \citep{bakke2023data, zhu2017size}. 

Finally, as satellite image time series are not evenly spaced due to cloud cover, we add as complementary information the day of the year for all our predictors composing the time series to inform the model on the image acquisition date. Other details on the experimental setup for all architectures can be found in \ref{sec:setup}. We analyze the impact of satellite image time series on model performance in \ref{sec:abl}.

\subsection{CNN-based Architecture}
\label{sec:cnn}

\begin{figure}[!t]
 \centering
 \makebox[\textwidth]{
    \resizebox{1.4\textwidth}{!}{
    \includegraphics[width=1.4\textwidth]{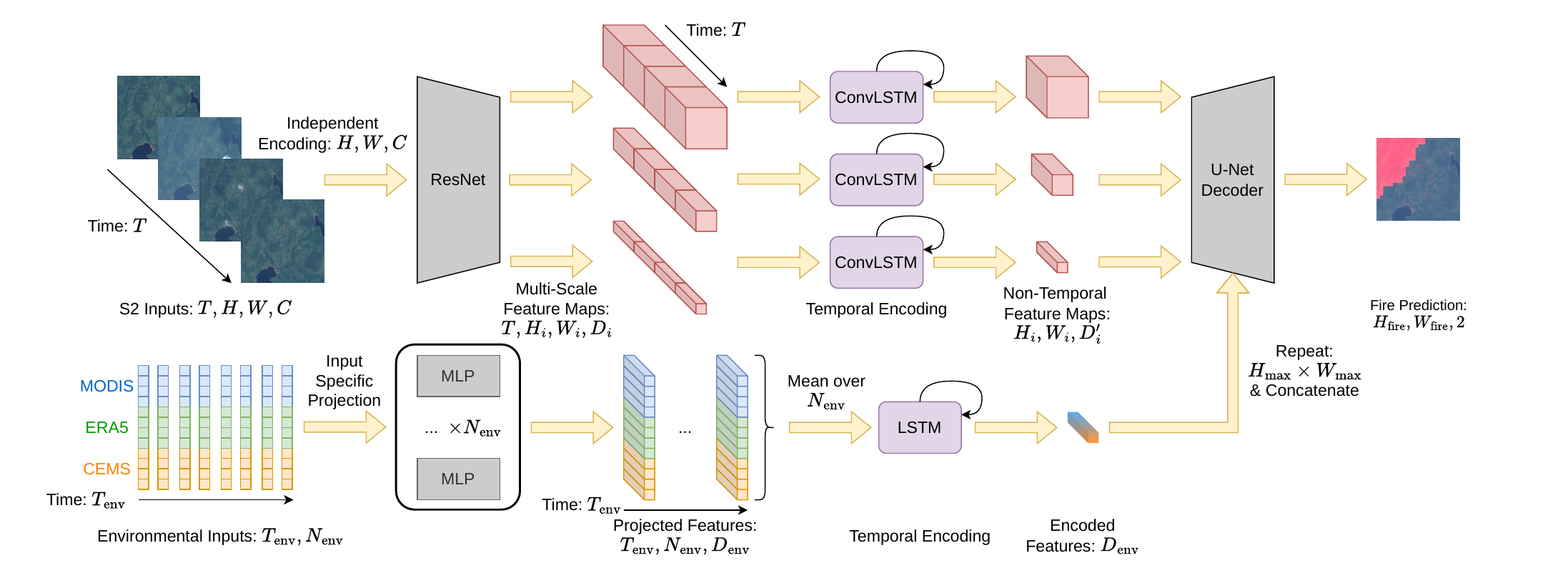}
    }
    }
  \caption{CNN Architecture for Wildfire Prediction in setting \Three. \textbf{Top:} Satellite image time series encoding, \textbf{Bottom:} Environmental predictors encoding.}
  \label{fig:cnn-multi}
\end{figure}

In the CNN-based architecture, satellite image time series are processed in a factorized manner: first spatially and then temporally, for both settings \One and \Three, as shown in Figure \ref{fig:cnn-img} and Figure \ref{fig:cnn-multi}, respectively. For a given satellite image time series $x_{1:T} = \{ x_t \}_{t=1}^{T}$, with $x_t \in \mathbb{R}^{H \times W \times C}$ being a single time step with $C$ the number bands and the day of the year, and $T$ a fixed number of time steps, each image $x_t$ is first encoded independently by a ResNet-50 pre-trained on ImageNet \citep{he2016deep}: $f(x_t) = \{ z_{i,t}\}_{i=1}^{N_S}$, with, $z_{i, t} \in \mathbb{R}^{H_i \times W_i \times D_i}$, which outputs $N_S=3$ feature maps of channel dimension $D_i$, each feature map corresponding to a different scale. The encoding of all time steps is done in parallel, and each scale-specific feature map, $z_{i, t}$, is concatenated independently for each scale across the temporal axis: $z_{i, 1:T} = \{z_{i, t}\}_{t=1}^{T}$. Then, the spatio-temporal encoding is done via one ConvLSTM model per scale. By extracting the last hidden state from each ConvLSTM: $g_i$, we obtain feature maps $g_i(z_{i, 1:T}) = s_i$, with $s_i \in \mathbb{R}^{H_i \times W_i \times D'_i}$ at 3 different scales with channel dimension $D'_i < D_i$, providing multiple levels of contextual information.

In setting \One (Figure \ref{fig:cnn-img}), our final multi-scale feature maps $\{s_i\}_{i=1}^{N_S}$ are passed to a U-Net-like decoder. The output of the decoder is interpolated to the dimensions of the label feature map: $H_{\text{fire}} = W_{\text{fire}} = \frac{H}{10} = \frac{W}{10}$ to match the patch resolution. This is finally passed to binary patch classification layer to output the class probabilities: $h({\{s_i\}_{i=1}^{N_S}}) = \hat{y} \in [ 0, 1]^{H_{\text{fire}} \times W_{\text{fire}} \times 2}$, with $h$ the function representing the decoder, interpolation, and patch classification layer.

In the multi-modal setting \Three (Figure \ref{fig:cnn-multi}), the above architecture is extended to process in parallel the time series of non-spatial environmental predictors
$x_{\text{env}, 1:T_{\text{env}}} = \{ x_{\text{env},t} \}_{t=1}^{T_{\text{env}}}$, with $ x_{\text{env}, t} \in \mathbb{R}^{N_{\text{env}}}$ being the data for a single time step with $N_{\text{env}}$ environmental predictors. $T_{\text{env}}$ is a fixed number of time steps. Following \citep{gorishniy2022embeddings} for tabular data encoding, each environmental predictor is projected to a high-dimensional space with specific MLP layers: $\forall j \in \{1, \dots, N_{\text{env}}\}, \; f_{\text{env}, j}(x_{\text{env}, t}^{j}) = z_{\text{env}, t}^{j}$, with $\; z_{\text{env}, t}^{j} \in \mathbb{R}^{D_{\text{env}}}$. The projected features are then averaged across the $N_{\text{env}}$ dimension to obtain $z_{\text{env},1:T_{\text{env}}} \in \mathbb{R}^{D_{\text{env}} \times T_{\text{env}}}$ and passed to an LSTM model for temporal encoding $g_{\text{env}}(z_{\text{env},1:T_{\text{env}}}) = s_{\text{env}} \in \mathbb{R}^{D_{\text{env}}}$ , which we use as the final environmental encoded feature. This one-dimensional vector is replicated spatially and concatenated with the final feature map from the U-Net-like decoder before the patch classification layer: $h({\{s_i\}_{i=1}^{N_S}}, s_{\text{env}}) = \hat{y} \in [ 0, 1]^{H_{fire} \times W_{fire} \times 2}$.

For setting \Two (Figure \ref{fig:cnn-env}), where the model is trained using only environmental predictors at a resolution varying from $500$ m up to $28$ km (see Table \ref{tab:predictors}), we first split the predictors into two groups: mid-resolution inputs $x_{\text{mid}, 1:T_{\text{env}}} = \{ x_{\text{mid},t} \}_{t=1}^{T_{\text{env}}}$, with single time step $x_{\text{mid}, t} \in \mathbb{R}^{H_{\text{mid}} \times W_{\text{mid}} \times N_{\text{mid}}}$ for all MODIS data, and low-resolution inputs for all ERA5 and CEMS data: $x_{\text{low}, 1:T_{\text{env}}} = \{ x_{\text{low},t} \}_{t=1}^{T_{\text{env}}}$, with single time step $x_{\text{low}, t} \in \mathbb{R}^{H_{\text{low}} \times W_{\text{low}} \times N_{\text{low}}}$. In this setting, we leverage spatial environmental inputs and not tabular to compensate for the absence of high-resolution satellite imagery from Sentinel-2, which provides spatial context for settings \One and \Three. In each group, as not all predictors have the same spatial resolution (see Table \ref{tab:predictors}), we upsampled all predictors to the highest available resolution. Details on the different spatial dimensions for each group can be found in the \ref{sec:setup}. We partially modify the architecture from setting \Three, as shown in Figure \ref{fig:cnn-env}. First, mid-resolution inputs are used as an alternative to satellite image time series. In practice, all the satellite image processing model components stay the same for the mid-resolution inputs group: the spatial encoding $f$, the scale-specific temporal encoding $g_i$, and the final head $h$ (corresponding to the decoder, interpolation layer, and patch classification layer). We simply extend the number of multi-scale feature maps to $N_S = 5$ because of the lower resolution of the input data. Moreover, we exchange ConvLSTM with LSTM when the output feature maps from $f$ become one-dimensional (for $i = 5$). The second branch of the model is adapted to process spatial data for the low-resolution inputs group. We use a smaller pre-trained CNN architecture to encode independently each time step, similarly to the processing of satellite images described above: we use ResNet-18 \citep{he2016deep} to obtain a one-dimensional feature vector $f_{\text{low}}(x_{\text{low}, t}) = z_{\text{low},t}$, with $z_{\text{low}, t} \in \mathbb{R}^{D_{\text{low}}}$. 
The temporally concatenated features $z_{\text{low},1:T_{\text{env}}} \in \mathbb{R}^{D_{\text{low}} \times T_{\text{env}}}$ are passed to an LSTM model $g_{\text{low}}(z_{\text{low},1:T_{\text{env}}}) = s_{\text{low}}$, with $s_{\text{low}} \in \mathbb{R}^{D'_{\text{low}}}$ to obtain low-resolution encoded features with $D'_{\text{low}} < D_{\text{low}}$. Similarly to the multi-modal architecture, this one-dimensional vector is replicated spatially and concatenated with the final feature map from the U-Net-like decoder that has processed the mid-resolution group.\\

In all three settings, the training is done with a per-patch loss, $L_{\text{CNN}}$, which is a combination of weighted cross-entropy loss and dice loss. Weighted cross-entropy gives more importance to the rare class \textit{fire} by increasing its contribution to the loss, while the dice loss measures overlap (i.e. intersection over union) and directly optimizes for better segmentation of small or imbalanced regions. The losses are as follows:

\begin{align}
    L_{\text{CNN}} &= L_{\text{WCE}} + L_{\text{DICE}} \: ,\\
    L_{\text{WCE}} &= - w_{\text{fire}} \sum_{i} y_i \log(\hat{y}_i) 
                     - w_{\text{no-fire}} \sum_{i} (1 - y_i) \log(1 - \hat{y}_i) \: ,\\
    L_{\text{DICE}} &= 1 - \frac{2 \sum_{i} y_i \hat{y}_i}{\sum_{i} y_i + \sum_{i} \hat{y}_i} \: ,
    \label{eq:dice}
\end{align}

where $ y_i $ is the ground truth label for a patch (1 for \textit{fire}, 0 for \textit{no fire}), $ \hat{y}_i $ is the predicted probability for fire, and $ w_{\text{fire}} $ and $ w_{\text{no-fire}} $ are class weights.

\subsection{Transformer-based Architecture}
\label{sec:vit}

\begin{figure}[t!]
 \centering
 \makebox[\textwidth]{
    \resizebox{1.4\textwidth}{!}{
    \includegraphics[width=0.1\textwidth]{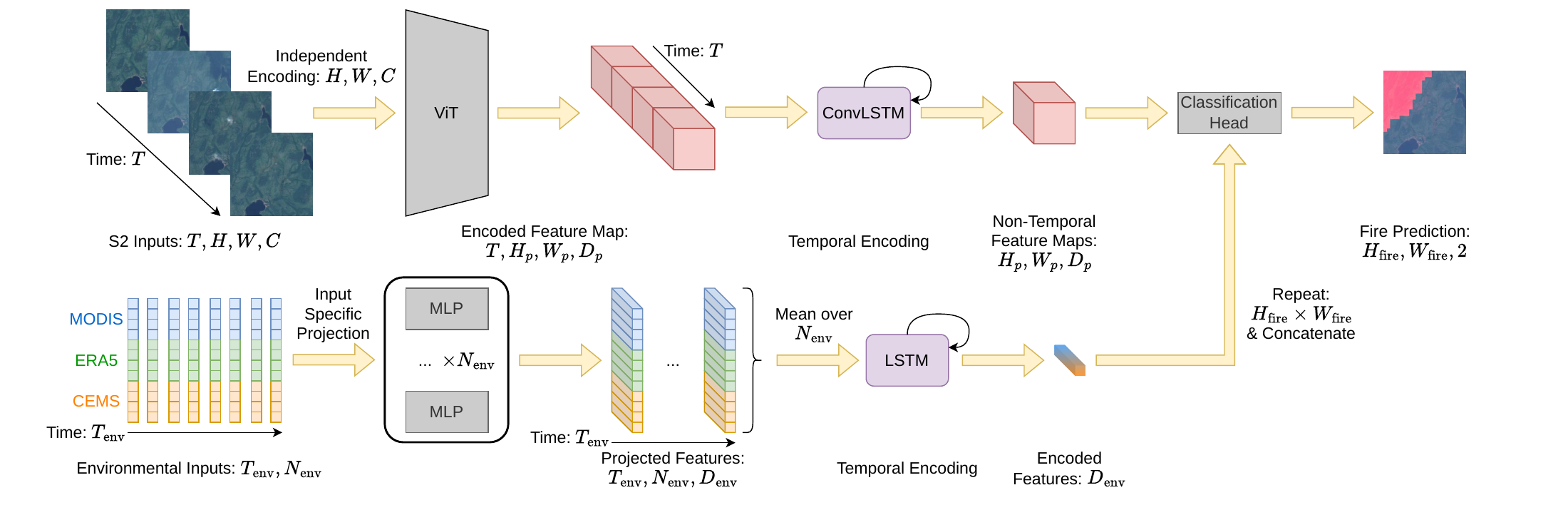}
    }
    }
  \caption{ViT Architecture for Wildfire Prediction in setting \Three. \textbf{Top:} Satellite image time series encoding, \textbf{Bottom:} Environmental predictors encoding.}
  \label{fig:vit-multi}
\end{figure}

In the three settings, our ViT architectures re-use most of the components of their CNN counterparts, as shown in Figure \ref{fig:vit-multi}, \ref{fig:vit-img}, and \ref{fig:vit-env}, respectively. The main difference is the absence of multi-scale feature maps after the satellite image encoding in options \One and \Three, or after the mid-resolution encoding for option \Two.

For a given satellite image time series $x_{1:T} = \{ x_t \}_{t=1}^{T}$, each image $x_t \in \mathbb{R}^{H \times W \times C}$ is encoded independently by a pre-trained ViT architecture, specifically DINOv2: ViT-S \citep{oquab2023dinov2}: $f(x_t) = \{ z_{t}\}$, with $z_{t} \in \mathbb{R}^{H_p \times W_p \times D_p}$, which outputs one feature map per time-step. Similarly to the CNN architecture, the encoding of all satellite images time steps is done in parallel, and the feature maps are concatenated across the temporal axis: $z_{1:T} = \{z_{t}\}_{t=1}^{T}$. As for the CNN, the temporal encoding is also done via a ConvLSTM model: $g(z_{1:T}) = s$, with $s \in \mathbb{R}^{H_p \times W_p \times D_p}$. Multi-scale feature maps are not necessary for ViT due to the native high-resolution of the final output feature map: $H_p$ and $W_p$. The output feature map, $s$, is interpolated to the label dimensions $H_{\text{fire}} = W_{\text{fire}} = \frac{H}{10} = \frac{W}{10}$ and finally passed to the model head, a patch classification layer, to output the class probabilities: $h_{\text{ViT}}(s) = \hat{y} \in [ 0, 1]^{H_{\text{fire}} \times W_{\text{fire}} \times 2}$, with $h_{\text{ViT}}$ the function representing the interpolation, and classification layer.

In the multi-modal model (\Three), the encoding of environmental inputs is identical to that of the CNN method. The final environmental encoded features $s_{\text{env}} \in \mathbb{R}^{D_{\text{env}}}$ is replicated spatially and concatenated with the final feature map $s$ before the patch classification layer to output the class probabilities: $h(s, s_{\text{env}}) = \hat{y} \in [ 0, 1]^{H_{\text{fire}} \times W_{\text{fire}} \times 2}$.

For option \Two, the same modifications from the satellite image time series are applied to the mid-resolution inputs; for low-resolution inputs, we use a ViT-S architecture similar to the one used for mid-resolution inputs, as it already represents the smallest available model for the DINOv2 architecture.

Contrary to the training for the CNN-based architectures, the loss used here is only the dice loss as defined in Equation \ref{eq:dice}, because experimentally it led to the best results.

\section{Results}
\label{sec:results}

This section details the key results for the benchmark models described in Section \ref{sec:methods}. CanadaFireSat covers the period 2016-2023: we train our models on the years 2016-2021, while keeping 2022 for validation (Val) and 2023 for both test sets: Test and Test Hard. 
Results are evaluated in terms of F1 score and PRAUC (Area Under the Precision-Recall Curve for the positive class \textit{fire} only). Both metrics are robust to imbalanced datasets, contrarily to patch-level accuracy. The F1 score is defined as the harmonic mean between Precision (proportion of true positive pixels over pixels predicted as positive) and Recall (proportion of true positives over all actual positives):
\begin{equation}
    \text{F1} = 2\times\frac{\textbf{Precision}\times\textbf{Recall}}{\textbf{Precision}+\textbf{Recall}} \: .
\end{equation}
It provides information about how well the model minimizes both false negatives and false positives at a fixed threshold. We favor the F1 score over Intersection over Union (IoU) as the former is more commonly used in the wildfire forecasting literature. PRAUC summarizes the Precision-Recall trade-off across all thresholds for the class \textit{fire}. PRAUC is used to compensate for the single arbitrary threshold of the F1 score. Both metrics are impacted by fire prevalence and showcase regional and inter-annual variability.

The benchmark models are tested against a baseline approach relying on the FWI in the following way: first, for a given time step $t$ we extract the 8-day mean FWI map at $0.25$° from $t - 8$ to $t - 1$ included
, and interpolate it at the target resolution of $100 \: \text{m} \times 100 \: \text{m}$, then, the per-patch prediction is obtained by binarizing the interpolated FWI map. The optimal threshold is tuned on the validation set, referring to the year 2022: $\text{FWI}_{\text{th}} = 6$. The PRAUC is computed by scaling the FWI values with respect to the maximum value: $\text{FWI}_{\text{max}} = 50$.

\subsection{Performance Analysis}
\label{sec:perf}

We evaluate the performance of the two different architectures (CNN and ViT)  across three different settings described in Section \ref{sec:methods}: \One, \Two, and \Three. The results are reported in Table \ref{tab:results}.

\begin{table}[b!]
    \centering
    \renewcommand{\arraystretch}{1.2}
    \resizebox{\textwidth}{!}{
    \begin{tabular}{llc|cccccc|cc}
        \toprule
        \multirow{2}{*}{\textbf{Encoder}} & \multirow{2}{*}{\textbf{Modality}} & \multirow{2}{*}{\textbf{Params (M)}} & \multicolumn{2}{c}{\textbf{Val}} & \multicolumn{2}{c}{\textbf{Test}} & \multicolumn{2}{c|}{\textbf{Test Hard}} & \multicolumn{2}{c}{\textbf{Avg}} \\
        \cmidrule(lr){4-5} \cmidrule(lr){6-7} \cmidrule(lr){8-9} \cmidrule(lr){10-11}
        & & & PRAUC & F1 & PRAUC & F1 & PRAUC & F1 & PRAUC & F1 \\
        \midrule
        \multirow{3}{*}{ResNet-50} & SITS Only & 52.2 & \underline{45.2} & \underline{49.3} & \underline{53.3} & \underline{58.9} & \underline{26.3} & \underline{36.7} & \underline{41.6} & \underline{48.3} \\
        & ENV Only & 97.5 & 41.6 & 46.7 & 49.9 & 53.5 & 24.5 & 33.1 &
        38.7 & 44.4 \\
        & Multi-Modal & 52.2 & \textbf{46.1} & \textbf{51.1} & \textbf{57.0} & \textbf{60.3} & \textbf{27.1} & \textbf{37.4} & \textbf{43.4} & \textbf{49.6} \\
        \midrule
        \multirow{3}{*}{ViT-S} & SITS Only & 36.5 & \textbf{45.2} & \textbf{50.6} & \underline{51.2} & 51.9 & \textbf{25.7} & 33.8 & \underline{40.7} & 45.2 \\
        & ENV Only & 54.8 & 34.8 & 45.7 & 49.2 & \textbf{59.9} & 21.2 & \underline{35.1} & 35.1 & \underline{46.9} \\
        & Multi-Modal & 37.7 & \underline{43.9} & \underline{50.0} & \textbf{56.3} & \underline{59.2} & \underline{25.1} & \textbf{36.6} & \textbf{41.8} & \textbf{48.6} \\
        \midrule
        Baseline (FWI) & ENV Only & - & 20.0 & 32.7 & 43.1 & 50.3 & 21.1 & 32.7 & 28.1 & 38.6 \\
        \makecell[l]{Baseline (UNet) \\ \citep{prapas2023televit}} & ENV Only & 9.1 & 33.6 & 43.2 & 51.4 & 58.4 & 25.1 & 34.2 & 36.7 & 45.3 \\
        \makecell[l]{Baseline (UTAE) \\ \citep{michail2025firecastnet}} & ENV Only & 1.1 & 32.9 & 43.8 & 47.2 & 52.5 & 22.0 & 31.7 & 34.0 & 42.7 \\
        \makecell[l]{Baseline (ConvLSTM) \\ \citep{yang2021predicting}} & SITS Only & 1.2 & 41.4 & 46.0 & 50.2 & 58.9 & 23.1 & 35.0 & 38.2 & 46.6 \\
        \bottomrule
    \end{tabular}}
    \caption{Performance comparison across different model settings and baselines. For our models, \textbf{bold} indicates the best metric value for each dataset split and model type, and \underline{underline} denotes the runner-up.}
    \label{tab:results}
\end{table}

 Across the three evaluation sets (last column of Table \ref{tab:results}), both ResNet-50 and ViT-S trained on \Three reach the highest performance, with $+15\%$ PRAUC and $+11\%$ F1 score for the CNN, compared to the FWI baseline. For both the CNN and ViT architectures, relying on multi-modal inputs shows, on average, an improvement over models trained on \One or \Two. While individually \One and \Two are already highly discriminative, the multi-modal setting \Three remains the most accurate forecaster with an average gain of $+1.8 \%$ in PRAUC and $+1.3 \%$ in F1 score for CNN-based models and $+1.1 \%$ in PRAUC and $+1.7 \%$ in F1 score for the ViT. We further discuss the role of satellite image time series and environmental predictors in Section \ref{sec:predictors}. On the Test set, the better performance seen in both the F1 score and PRAUC of all models set can be explained by fire patterns likely being more easily distinguishable due to the extreme fire season and by the relative increase in fire prevalence (comparatively to the Val set), a behavior also observed in the FWI baseline. Nonetheless, our best performing CNN model relying on \Three still outperforms the FWI baseline, on the Test set by $+13.9 \%$ (PRAUC) and by $+10 \%$ (F1 score). Further analysis of the model's performance threshold analysis across the Val, Test, and Test Hard sets is shown in \ref{sec:app-test}. The drop in performance of all models on the Test Hard set demonstrates the impact of the sampling strategy and the necessity of such an evaluation set. Test Hard can be used to assess models' lower bound performance and their ability to model the hidden phenomena behind ignition. In \ref{sec:proxy}, we show that simply extending the set of low- and mid-resolution predictors with ignition proxies only provides marginal performance gains in setup \Three, where we do not specifically train on hard samples. When it comes to the comparison between ResNet-50 and ViT-S trained on \Three, the former shows to perform best in terms of F1 score across all sets. However, differences remain small, and both architectures seem valid solutions for wildfire forecasting. Moreover, both ResNet-50 and ViT-S trained on \Three outperform all the deep learning based baselines trained on \One and \Two.

\begin{figure}[t!]
 \captionsetup[subfigure]{justification=centering}
  \centering
  \begin{subfigure}{0.48\textwidth}
    \centering
    \includegraphics[width=\textwidth]{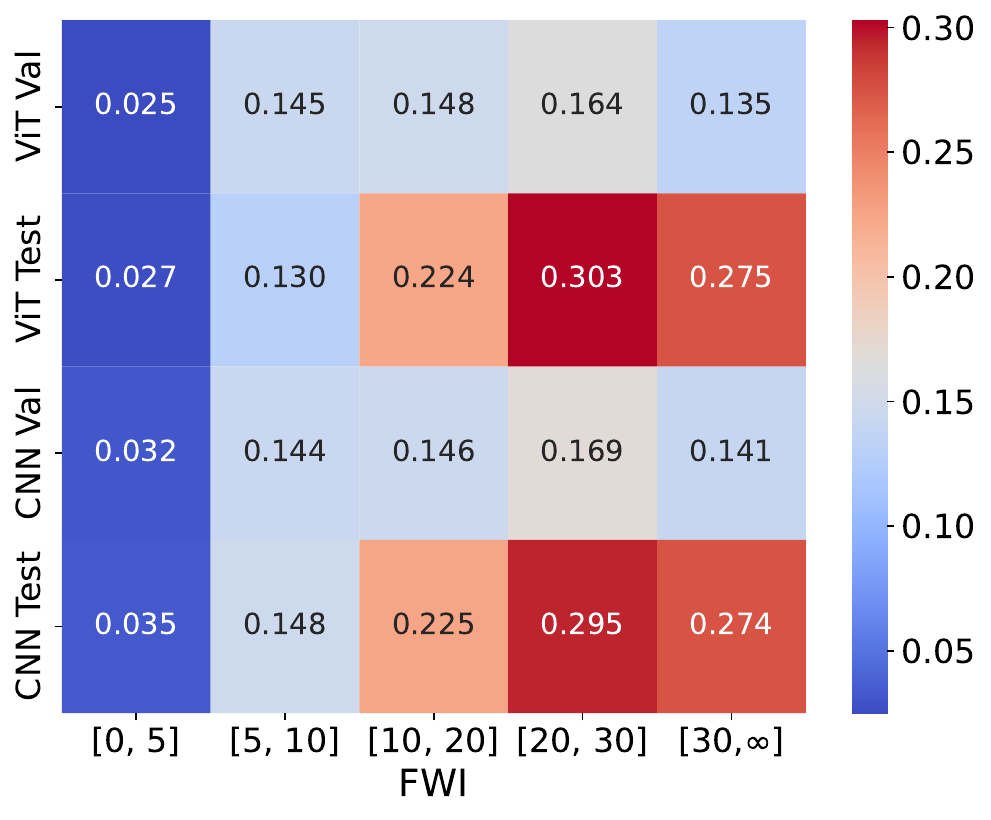}
    \subcaption{Negative samples FPR for the Multi-Modal methods across FWI levels.}
    \label{fig:fpr-h}
  \end{subfigure}
  \hfill
  \begin{subfigure}{0.48\textwidth}
    \centering
    \includegraphics[width=\textwidth]{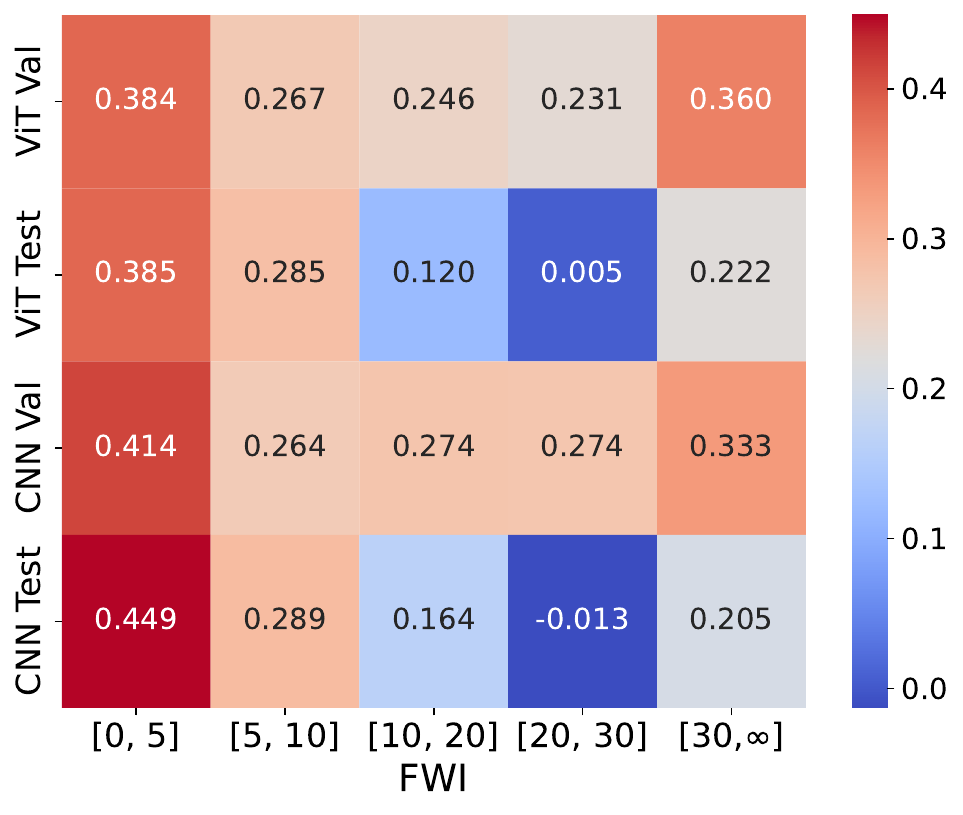}
    \subcaption{Weighted F1 score for the Multi-Modal methods across FWI levels.}
    \label{fig:f1-h}
  \end{subfigure}
  \caption{\Three models' performance across different FWI value groups.}
  \label{fig:fwi-perf}
\end{figure}

The performances of the different models in setting \Three are studied in Figure \ref{fig:fwi-perf} for increasing FWI values. We first focus on the False Positive Rate, defined as $\text{FPR} = \frac{\text{FP}}{\text{FP} + \text{TN}}$. As expected, we can observe in Figure \ref{fig:fpr-h} a positive correlation between the FPR and the FWI. Indeed, negative samples associated with a higher FWI show similar fire danger conditions to positive samples, and are thus much more difficult to discriminate, with ignition becoming the main triggering factor for samples with  $\text{FWI} > 20$. 
Then, we study in Figure \ref{fig:f1-h} the variations of the weighted F1 score, defined as $\hat{\text{F1}} = \frac{\text{F1} - \text{F1}_{\text{pos}}}{1 - \text{F1}_{\text{pos}}}$. This second index tells how good the model is compared to a naive predictor: $\text{F1}_{\text{pos}}$, assigning the class \textit{fire} to all samples. We note a negative correlation between $\hat{\text{F1}}$ and the FWI: as the FWI increases, there is approximately no difference between our benchmark models and a naive predictor. Such behavior is not surprising as it is more likely for a sample associated to a high FWI to belong to the class \textit{fire}, as confirmed by the increase in percentage of positive samples with higher FWI (from $13\%$ at $\text{FWI} \in [0, 5]$ to $77 \%$ at $\text{FWI} \in [20, 30])$. We can conclude that the improved performance of our model with respect to the FWI baseline reported in Table \ref{tab:results} is due to a better prediction of wildfire occurrence at lower FWIs, as the task becomes trivial for $\text{FWI} > 20$ due to data imbalance.

\begin{figure}[t!]
 \centering
    \includegraphics[width=0.48\textwidth]{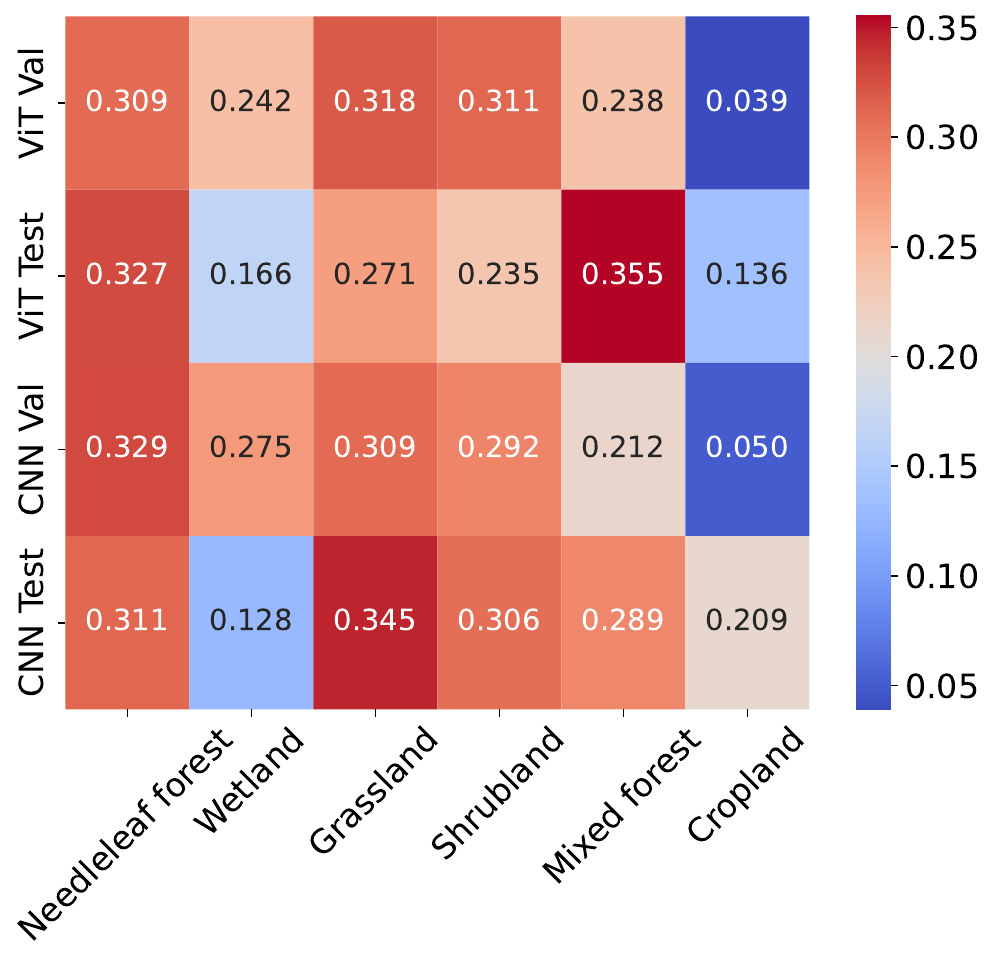}
  \caption{\Three models' F1 score across the main land cover classes.}
  \label{fig:lc-h}
\end{figure}

We also study the results of ResNet-50 and ViT-S trained on \Three across the most common land cover classes in CanadaFireSat (see Figure \ref{fig:lc-h}). Our models struggle the most on the classes \textit{wetland} and \textit{cropland}. Indeed, fire patterns in these two land cover types differ from those observed in the majority of wildfires, which tend to affect forest areas. In particular, peatland fires in Canada can occur under the ground in wet areas, or even under the snow layer. Such fires are difficult to observe through the predictors considered in the proposed CanadaFireSat, and would require \textit{ad-hoc} modeling due to the specificities of such ecosystems. Low scores are also observed for cropland fires, which also present unique fire patterns, as the ignition is often human-induced and driven by a specific need for agricultural practices. As before, detecting these events seems hardly possible with our remote sensing-based system.

\subsection{Deployment at Scale: Case Study}

CanadaFireSat enables training deep models for high-resolution wildfire forecasting. As a result, our dataset makes it possible to deploy models capable of monitoring large regions at high-resolution. In this section, we demonstrate on a real use case how a model trained on our CanadaFireSat dataset could be deployed at a scale useful for wildfire management teams.

\begin{figure}[htbp]
\centering
\resizebox{\columnwidth}{!}{
 \begin{tabular}{lc}
 \rotatebox{90}{\parbox{6cm}{\centering Original S-2 Tile}} & 
 \includegraphics[width=\textwidth, height=6cm]{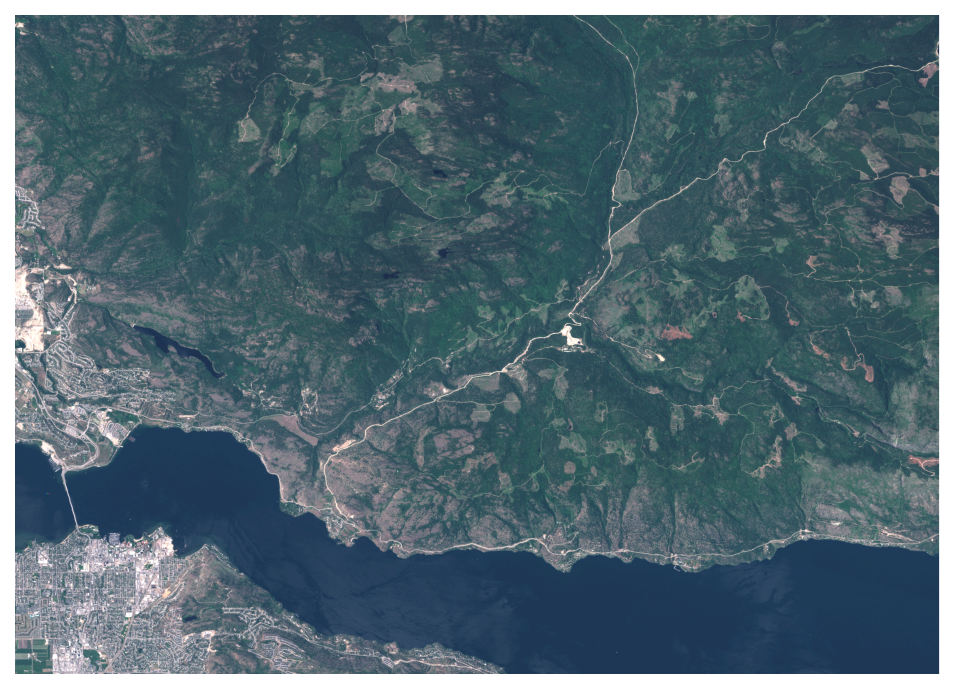} \\
 \rotatebox{90}{\parbox{6cm}{\centering Target Fire Polygon}} & 
 \includegraphics[width=\textwidth, height=6cm]{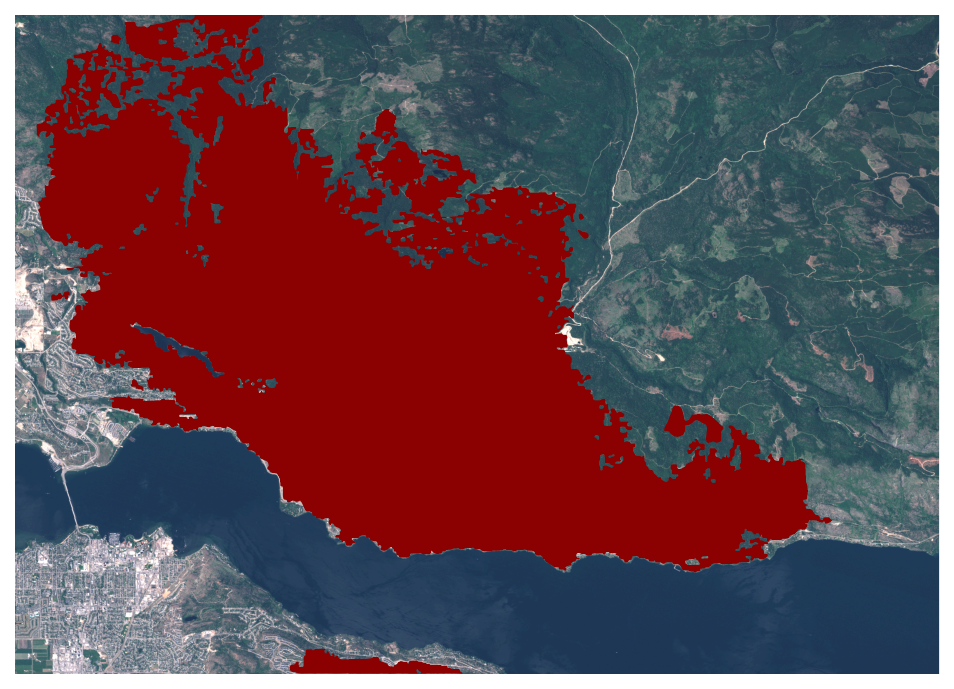} \\
 \rotatebox{90}{\parbox{6cm}{\centering Model Predictions}} & 
 \includegraphics[width=\textwidth, height=6cm]{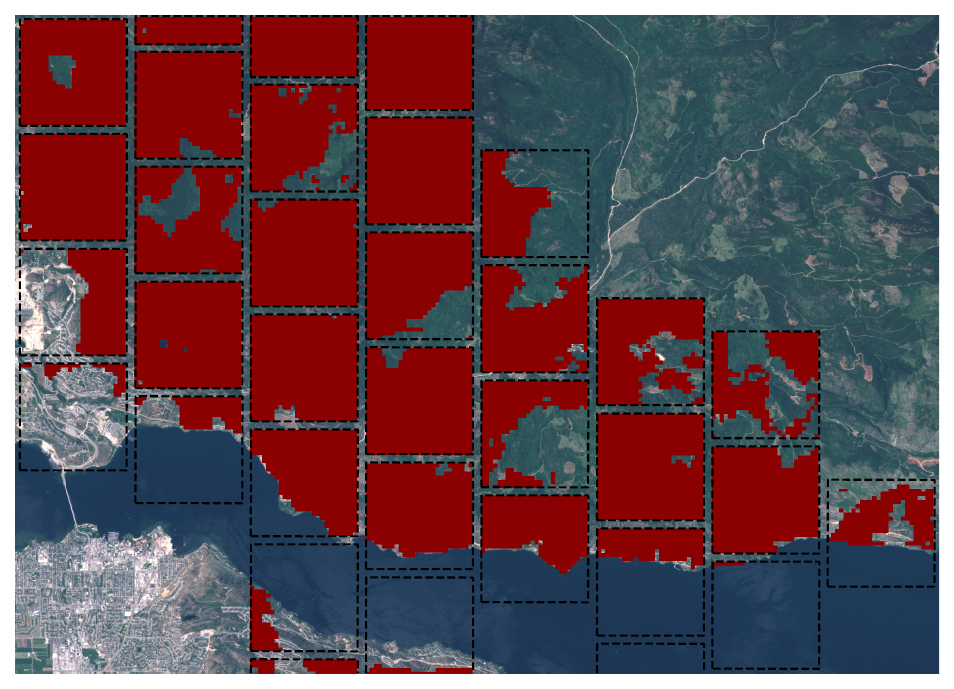}
 \end{tabular}}
 \caption{\textbf{Row 1} Sentinel-2 tile from $2023/06/06$ of size $16 \: \text{km} \times 22 \: \text{km}$ before a large wildfire in British Columbia. \textbf{Row 2} Fire polygons for the large wildfire on $2023/07/01$ over the same tile. \textbf{Row 3} Binary model predictions (in \textcolor{red}{\textbf{red}}) over the $2.64 \: \text{km} \times 2.64 \: \text{km}$ center-cropped positive samples (patches outlined in \textbf{black}).}
 \label{fig:use case}
\end{figure}

We chose as a case study a large wildfire that occurred in British Columbia on $2023/07/01$, illustrated in Figure \ref{fig:use case}. The first row displays the RGB composite of the region of interest of size $16 \: \text{km} \times 22 \: \text{km}$ acquired by Sentinel-2 right before the wildfire starts on $2023/06/06$. The fire scar polygons from NBAC are shown on the second row. The third row shows the binarized predictions of the CNN model trained in the multi-modal setting \Three, on the positive samples overlapping with the considered ROI. We observe how well the model delineates the urban interface on the left side of the wildfire and the rough approximation of its boundaries on the right side of the fire. However, we can also see at the top of the Sentinel-2 image that the model overestimates the wildfire extent. This case study showcases the potential of CanadaFireSat to enable the deployment of models capable of monitoring large regions at the unprecedented resolution of $100$ m.

\section{Discussion}

\subsection{High-resolution Wildfire Forecasting via Multi-modal Learning}
\label{sec:predictors}

As previously demonstrated in \citep{pelletier2023wildfire, chowdhury2015operational, yang2021predicting}, multi-spectral multi-temporal satellite data can be a valuable data source to forecast wildfires. Indeed, several spectral indices discriminative for wildfire forecasting can be extracted from Sentinel-2: normalized difference vegetation index (NDVI), normalized difference water index (NDWI), tasseled cap wetness, and channel histograms. The results reported in Table \ref{tab:results} demonstrate the potential of multi-spectral temporal satellite data for high-resolution wildfire forecasting (setting \One). While hydrometeorological data (setting \Two) are commonly used in global and continental wildfire forecasting models, they can be complemented by satellite data to improve strongly both the spatial resolution and the accuracy of the prediction (setting \Three), as reported in Table \ref{tab:results}, where the multi-modal approach leads to the best performances. 

\subsection{Importance of Negative Sampling for Training and Evaluation}
\label{sec:dis-samp}

Numerous wildfire forecasting benchmarks require sampling the negative (non-fire) samples due to extreme imbalance and computational constraints. A common strategy is to focus on samples that burned once across the period studied \citep{bakke2023data, prapas2022deep, prapas2023televit}. In CanadaFireSat, we opt for a different strategy: we sample negative examples for each Canadian province uniformly across their yearly fire driver patterns (FWI values). By sampling the training, validation, and test sets in this way, we aim to train and evaluate our models on a subset representative of the conditions encountered in all of Canada. Nonetheless, as the yearly fire patterns vary, the distribution of negatives with respect to the FWI changes over the years, in turn affecting the performance of models. This motivated the creation of a second test set to understand the impact of sampling on the models' performance (Test Hard), 
where ignition, a complex phenomenon difficult to model \citep{chen2021future, calef2008human}, differentiates positive from negative patches. Indeed 
human-induced ignitions, generally caused by infrastructures, agricultural practices, or "recreational" activities, are typically hard to estimate with CanadaFireSat, as Sentinel-2 is the only source providing information on human presence, but only for a limited spatial context of $2.64 \: \text{km} \times 2.64 \: \text{km}$. 
Fine-tuning our multi-modal models on data such as Test Hard, while also enhancing our set of predictors with proxies of ignition probability (as in \ref{sec:proxy}), is a relevant direction for improving our models towards accounting for ignition probability.

\subsection{Modeling Wildfires in the Boreal Ecosystem}

As initially stated in Section \ref{introduction}, one of our main motivations is the rise of wildfires in the boreal ecosystem and the risks this represents for its local communities. To cover the areas of interest and to evaluate the broader impact of wildfires on global climate, we created our benchmark CanadaFireSat so that it covers the entirety of Canada, including all its agricultural lands, urban areas, and other ecosystems such as the temperate forest in British Columbia. To study the behavior of trained models on the boreal ecosystem, it is possible to constrain the analysis on the main land cover classes of the boreal ecosystem (needleleaf forest and wetlands). With our benchmark models trained on CanadaFireSat, we observe an important difference of performance between those two land cover classes: with Multi-Modal CNN and ViT performing respectively $+ 11.9 \%$ and $+ 11.4 \%$ better on needleleaf forest compared to wetlands in terms of F1 score on both the Val and Test sets, showing that for the latter land cover, performance is still not optimal. Indeed, wetland wildfires are a unique phenomenon compared to forest wildfires, as they depend much more on soil-related predictors and can burn underground for a long period. As a consequence, they are sometimes undetectable for optical remote sensing satellites. In particular, peatland wildfires that emit large amounts of CO2 and mercury \citep{fraser2018important, kohlenberg2018controls} are commonly studied independently from forest fires \citep{pelletier2023wildfire, bali2021prediction}. Extending CanadaFireSat so that it includes data acquired from radar remote sensing satellites (for instance, Sentinel-1 images) could help to better model the surface soil conditions for wetlands \citep{millard2018quantifying} and bridge the gap in performance across the boreal ecosystem.

\subsection{Operationalization of the Model}
\label{sec:dis-op}

Deploying models trained with CanadaFireSat over the entirety of Canada would require densely sampling the country with Sentinel-2 image time series, resulting in a huge amount of data to be processed. Indeed, the proposed dataset is aimed at modeling wildfire patterns at a moderate scale, but at high-resolution, and can be coupled with coarser resolution approaches \citep{prapas2022deep, bali2021prediction} to identify areas of interest and then apply our model to map such areas more precisely. Such coupling would allow wildfire management experts to target specific areas at risk for fine-grained wildfire forecasting or focus on areas that require more surveillance due to their proximity to local communities or due to their ecological and environmental interest. By alleviating the need for significant computational resources, it would break the barrier to scale this approach to large continental areas. Learning models directly capable of multi-scale prediction is an interesting future research direction to deploy high-resolution wildfire forecasting at scale. Such a model would exploit hierarchical learning approaches developed in computer vision for semantic segmentation \citep{li2022deep, atigh2022hyperbolic}. Operationalization of the method might also require means of evaluation more closely related to wildfire management objectives, such as controlling vegetation and infrastructure damage or dangerous emissions. 

\subsection{Limitations and Future Work}

As mentioned in Section \ref{sec:dis-samp}, the main limitation of methods trained on CanadaFireSat is the difficulty of modeling the ignition component in wildfires due to its inherent stochasticity. Weather data from ERA5 can provide information on the risk of lightning; nonetheless, explicitly adding lightning probability \citep{geng2019lightnet} as a predictor, as well as training on a subset of samples with high fire weather risk and lightning occurrence, could help the trained models to better characterize ignition.

Multi-task learning \citep{zhang2021survey} with multi-scale and multi-view representations \citep{wang2020deep, liu2021swin} could also be leveraged to develop a model forecasting wildfires at multiple scales. One could leverage different forecasting heads at multiple resolutions: $10$ km, $1$ km, $100$ m. This could help alleviate memory size constraints when high-resolution forecasts are deemed unnecessary and help provide consistent predictions across scales, and therefore alleviate continuity breaks across neighboring tiles, as shown in Figure \ref{fig:use case}.

Moreover, one could investigate the potential of geolocation embeddings such as SatCLIP \citep{klemmer2023satclip} or GeoCLIP \citep{vivanco2023geoclip} to represent high-resolution non-dynamic satellite information. These could be combined with non-spatial, but temporal dynamics from Sentinel-2 \citep{pelletier2023wildfire} as a way to factorize spatial and temporal components in satellite data and limit memory consumption. Extending CanadaFireSat with atmospherically corrected images (e.g. L2A) or with BRDF-corrected Harmonized Landsat and Sentinel-2 data could help improving performances. 

Another line of future research deals with the improvement of the pretraining of our multi-modal deep learning approaches. In our work, we leverage image encoders pre-trained on natural images such as ImageNet or via DINOv2, which are very different from multi-spectral satellite images. With the drastic increase in availability of Earth observation data, several models are being proposed to learn in an unsupervised way generalizable representations from this data \citep{cong2022satmae, jakubik2023foundation, hong2024spectralgpt, astruc2024anysat, sumbul2025smarties}. 
One could study the potential of those foundation models as pre-trained representations to be used in high-resolution wildfire forecasting; CanadaFireSat could be the perfect starting point for such an investigation. Moreover, as weather and climate data are currently omitted from Earth observation foundation models, future works could focus on fusion strategies that synergize environmental predictors with those general embeddings. Also, the stage where the fusion happens is worth investigation: in our method, we explore late fusion, but other strategies exist \citep{hong2020more}, with commonly used early-mid fusion strategies \citep{astruc2024anysat, sumbul2025smarties, jakubik2025terramind}. Thanks to the multiple attention types~\citep{hendricks2021decoupling}, the transformer architecture enables multiple types of fusion at different stages of the model.

Deep learning pipelines are affected by two sources of uncertainty: those coming from the data and those coming from the model. Regarding the first, future works could focus on the pre-processing of L1C products tailored for wildfire forecasting. Additionally, one could focus on the data gaps in the input time series: while environmental time series are always complete, SITS can be incomplete due to cloud cover. This factor is mitigated by temporal-aware modeling similar to what we proposed (the day of the year associated with each Sentinel-2 image is passed to the model), making the model robust to data gaps in the input SITS (see \ref{sec:cloud}). Data uncertainty can be further mitigated through cloud removal strategies \citep{zhang2020thick, wright2024clouds2mask, wright2025training}. Moreover, while setting \Two already shows good performance at $100$ m resolution, the use of high-resolution SITS further reduces the uncertainty at the target resolution of $100$ m.  Regarding the second (model uncertainty), hyperparameters have been tuned on an independent validation set, which has proven to be a relevant test bed for the choice of the best set of hyperparameters. Furthermore, we release two test sets corresponding to different fire regimes, which can be used in follow-up studies to assess model uncertainty in extreme wildfire seasons.


Finally, the increased complexity of models raises concerns regarding their interpretability and the possibility of understanding the role of the input variables in the final predictions \citep{ejaz2025comprehensive}. Several approaches exist to provide interpretations of black box wildfire forecasting models via feature attributions \citep{sundararajan2017axiomatic, selvaraju2017grad} or ranking \citep{lundberg2017unified}, or even to directly build interpretable wildfire forecasting model architectures \citep{koh2020concept, chen2019looks} via dense prediction architecture \citep{sacha2023protoseg, porta2025multi}. However, those methods need adaptation to accommodate multi-modal \citep{ekim2023explaining, wang2023visual} or multi-temporal data \citep{turbe2023evaluation, gee2019explaining}. They are also often not directly applicable to Earth observation data \citep{porta2025interpretable} due to their strong implicit bias for natural images \citep{chen2019looks}. This gap remains unfulfilled, and future works, for and beyond the wildfires prediction problem, should explore interpretable methods specifically tailored to Earth observation problems.

\section{Conclusion}

In this paper, we introduced CanadaFireSat, a comprehensive benchmark dataset for high-resolution wildfire forecasting over Canada from 2016 to 2023. CanadaFireSat was constructed to support multiple settings for model training: \One, \Two, and \Three. We demonstrated experimentally the potential of multi-modal learning for high-resolution wildfire forecasting on CanadaFireSat across two architectures: ResNet and ViT. Moreover, our experiments showed the importance of negative sampling in the evaluation of wildfire forecasting models. CanadaFireSat aims to accelerate research towards high-resolution monitoring of at-risk regions of interest to support wildfire management teams who are tasked with monitoring and protecting vast areas, such as the boreal ecosystem covering much of Canada. Results from this work demonstrate the feasibility of constructing future datasets like CanadaFireSat for other fire-prone landscapes where high-resolution fire polygons are available, like the Pan-Arctic, Pan-boreal, and grassland and forest ecosystems of the Tropics, since all input variables are globally available and open-access, even though certain fire regimes might require other high-resolution sensors, as seen for peatland fires. We hope this open-access dataset will foster research in this direction. 



\clearpage

\section{Data Availability}

CanadaFiresat is directly available on the \href{https://huggingface.co/datasets/EPFL-ECEO/CanadaFireSat}{Hugging Face Hub}, and the code for the \href{https://github.com/eceo-epfl/CanadaFireSat-Data}{data generation} and the \href{https://github.com/eceo-epfl/CanadaFireSat-Model}{model benchmarking} is open-access.

\section{Acknowledgments}
Devis Tuia was supported by the Horizon Europe grant 101213369 (DVPS) and the State Secretariat for Education, Research and Innovation (contract 25.00150).
\bibliographystyle{plainnat}
\bibliography{egbib}



\newpage
\appendix

\section{Illustration of the Environmental Predictor Alignments}
\label{sec:align}

\begin{figure}[htpb]
    \centering
    \includegraphics[width=\textwidth]{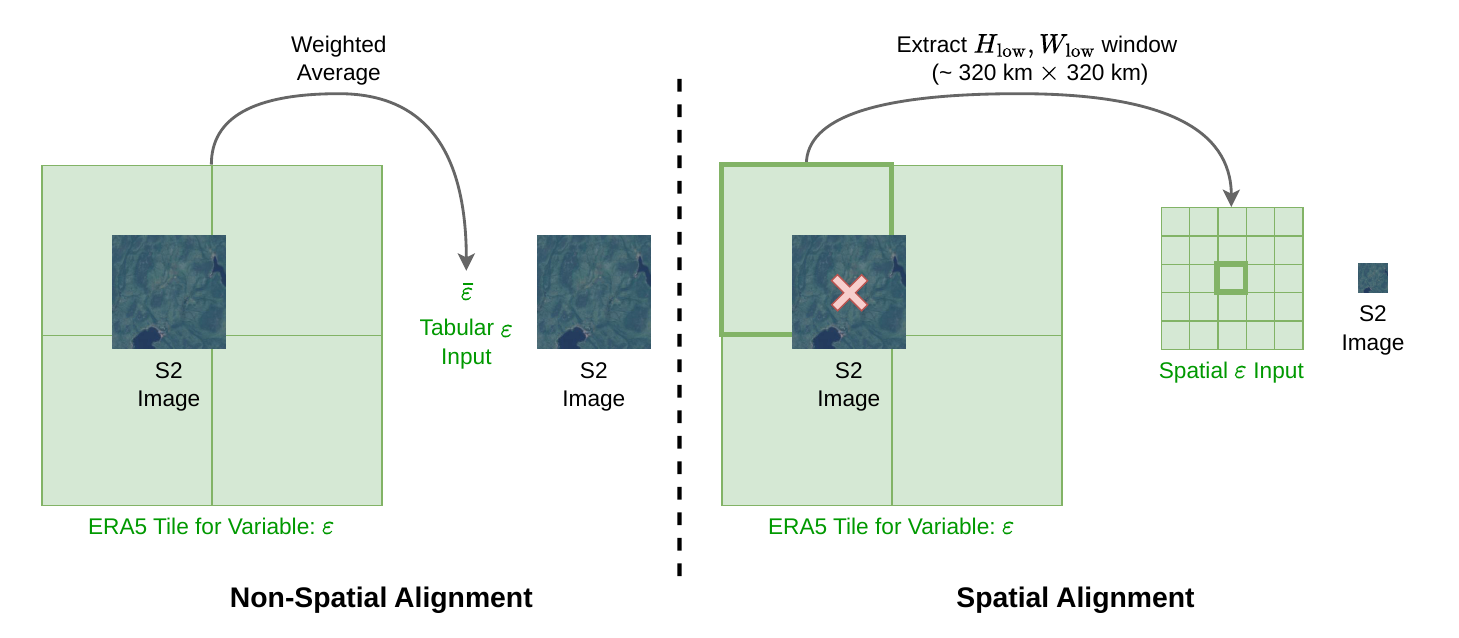}
    \caption{Illustration for ERA5 of the non-spatial (left) vs spatial alignment (right) of the environmental predictors.}
    \label{fig:align}
\end{figure}

In Figure~\ref{fig:align}, we illustrate the two different processes to extract the environmental predictors for the setups \Two and \Three.

\section{Model Architectures}
\label{sec:model-arch}

\begin{figure}[h!]
 \centering
 \makebox[\textwidth]{
    \resizebox{1.4\textwidth}{!}{
    \includegraphics[width=1.4\textwidth]{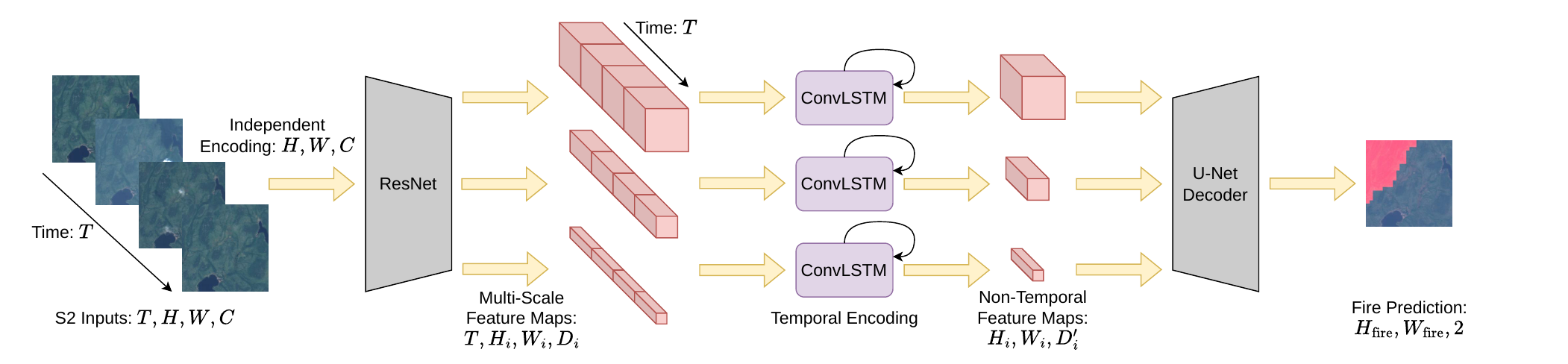}
    }
    }
  \caption{CNN Architecture for Wildfire Prediction used for setting \One.}
  \label{fig:cnn-img}
\end{figure}

\begin{figure}[h!]
 \centering
 \makebox[\textwidth]{
    \resizebox{1.4\textwidth}{!}{
    \includegraphics[width=1.4\textwidth]{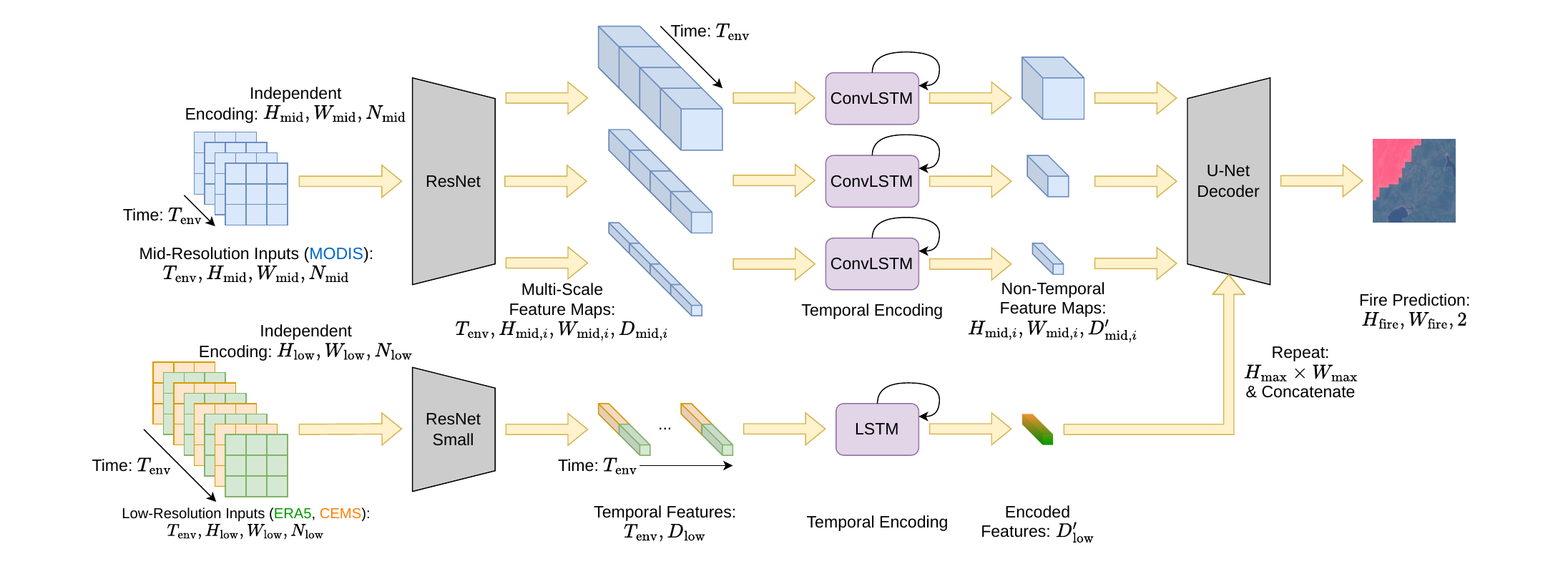}
    }
    }
  \caption{CNN Architecture for Wildfire Prediction used for Setting \Two.}
  \label{fig:cnn-env}
\end{figure}

\begin{figure}[h!]
 \centering
 \makebox[\textwidth]{
    \resizebox{1.4\textwidth}{!}{
    \includegraphics[width=1.4\textwidth]{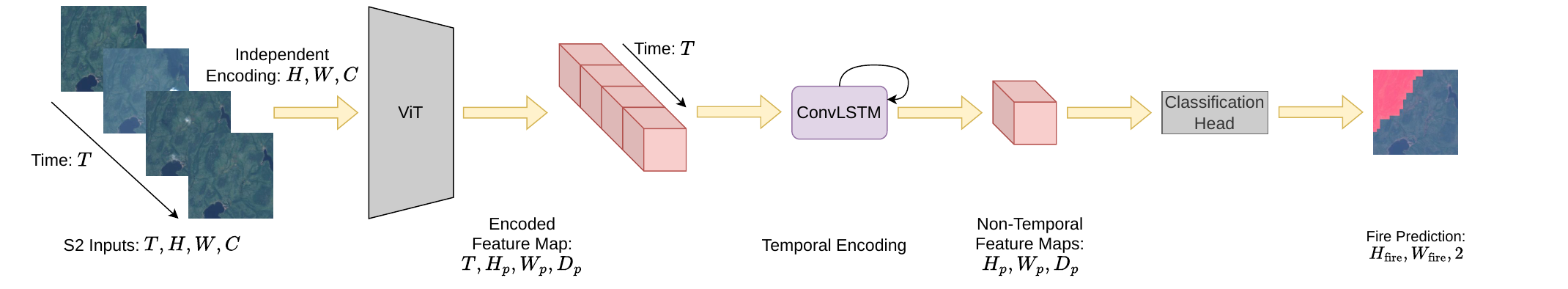}
    }
    }
  \caption{ViT Architecture for Wildfire Prediction used for Setting \One.}
  \label{fig:vit-img}
\end{figure}

\begin{figure}[h!]
 \centering
 \makebox[\textwidth]{
    \resizebox{1.4\textwidth}{!}{
    \includegraphics[width=1.4\textwidth]{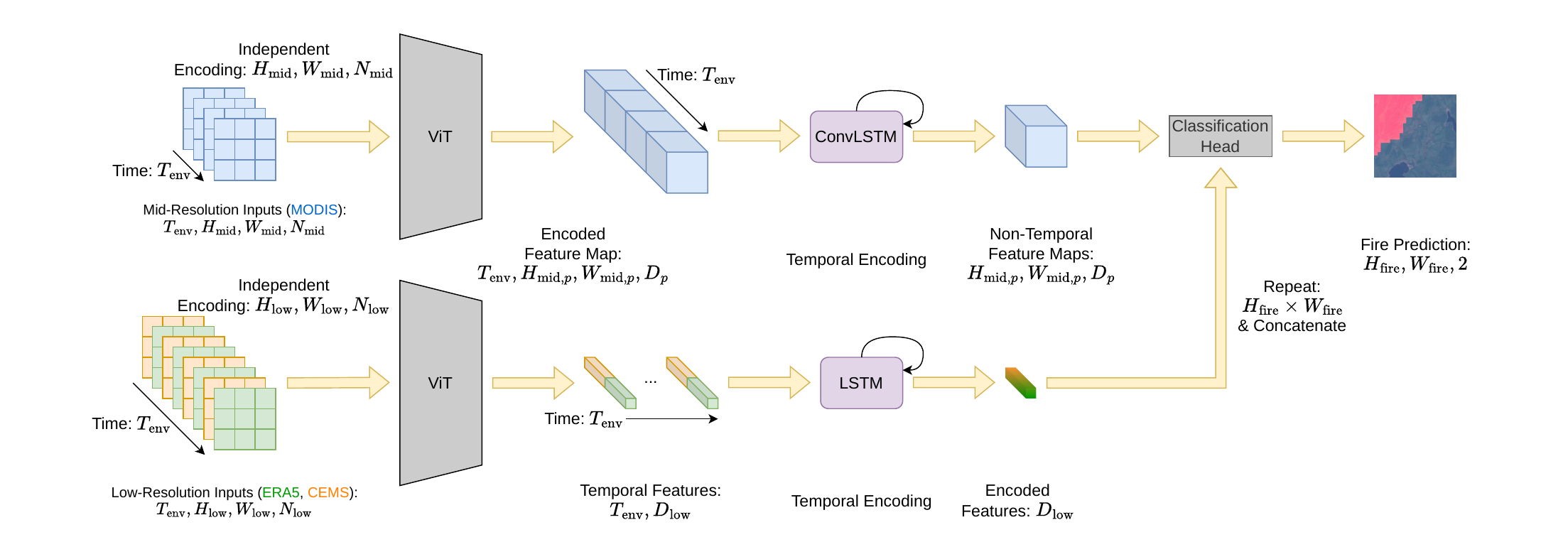}
    }
    }
  \caption{ViT Architecture for Wildfire Prediction used for Setting \Two.}
  \label{fig:vit-env}
\end{figure}

In this section, we illustrate the architectures used in the settings \One (Figure \ref{fig:cnn-img}) and \Three (Figure \ref{fig:cnn-env}) for the CNN-based models. We then show those used in the Transformer-based models in Figures \ref{fig:vit-img} and \ref{fig:vit-env}, respectively.

\section{Experimental Setup:}
\label{sec:setup}

\subsection{CNN Architecture Parameters}

As mentioned in Section \ref{sec:cnn}, satellite image time series encoding is done via a ResNet-50 backbone pre-trained on ImageNet. During training, the inputs are of size $T = 5$, $C = 14$, and $H = W = 240$ leading to a target resolution $H_{\text{fire}} = W_{\text{fire}} = 24$. During testing, we compute the prediction on the whole sample of size $H = W = 260$ with the same $T$ and $C$, leading to $H_{\text{fire}} = W_{\text{fire}} = 26$. We extract the model's last three feature maps of channel dimensions: 512, 1024, and 2048. Those feature maps pass through three independent ConvLSTM models, each one with kernel size $3 \times 3$ and only one layer. The ConvLSTM models output feature maps of dimensions: 64, 128, and 256, which are then passed to a U-Net-like decoder and interpolated to the target size.

This model is extended to multi-modal inputs of dimensions $N_{\text{env}} = 15$, including the day of the year, and $T_{\text{env}} = 8$, as we leverage the whole time series for those inputs. This data is projected to a high-dimensional space of size $D_{\text{env}} = 64$ and passed to an LSTM with one layer. The selected environmental predictors are the following: Total Precipitation Sum: 8-day Mean, Skin Temperature: 8-day Mean, Temperature (2m): 8-day Mean, Volumetric Soil Water Layer 1: 8-day Mean, Wind Speed (10m): 8-day Mean, Relative Humidity: 8-day Mean, Vapor Pressure Deficit: 8-day Mean, LST Day (1km): 8-day Mean, NDVI, EVI, FPAR, LAI, Drought Code: 8-day Mean, Fire Weather Index: 8-day Mean.

The model using only environmental predictors leverages inputs of dimension $H_{\text{mid}} = W_{\text{mid}} = 32$ for mid-resolution data (MODIS), and  $H_{\text{low}} = W_{\text{low}} = 32$ for low-resolution data (ERA5, CEMS). MODIS data at $1$ km: LST Day is interpolated to $500$ m to align with the rest of the MODIS inputs. Similarly, for the CEMS data, the Fire Weather Index and Drought Code, originally at $0.25$°, are interpolated to $11.1$ km to align with ERA5-Land. We leverage the same set of environmental predictors as in the multi-modal setting, split into the two resolution groups. The temporal dimension of those inputs is $T_{\text{env}} = 8$, the number of mid-resolution predictors is $N_{\text{mid}} = 6$ including the day of the year, and the number of low-resolution predictors is $N_{\text{low}} = 10$ including the day of the year (one more dimension than $N_{\text{env}}$, as we include the day of the year twice). As mentioned in Section \ref{sec:cnn}, for the mid-resolution group we leverage $N_S = 5$ multi-scale feature maps of dimensions: $64$, $256$, $512$, $1024$, and $2048$, for the mid-resolution data. The last feature map of channel dimension $2048$ is one-dimensional and is passed to an LSTM network for the temporal encoding. For the other four, we use independent ConvLSTM models. Those temporal encoders output feature maps of dimensions $64$, $128$, $256$, $512$, and $1024$, which are passed to a U-Net-like decoder and interpolated to the target size. The low-resolution inputs are encoded via a smaller network: ResNet-18, which outputs feature maps of channel dimension $D_{\text{low}} = 512$, encoded temporally with a LSTM with of one layer to a dimension $D'_{\text{low}} = 64$, matching the channel dimension of the last feature map of the U-Net decoder. Both ResNet encoders are pre-trained on ImageNet. As for the other settings, the training is done with a target resolution of size  $H_{\text{fire}} = W_{\text{fire}} = 24$, and at test time the target resolution is $H_{\text{fire}} = W_{\text{fire}} = 26$.

\subsection{ViT Architecture Parameters}

In the case of the ViT architecture, the satellite image time series encoding is done via the DINOv2 ViT-S architecture. The input channel and temporal dimensions are the same as for the CNN architecture: $T = 5$, and $C = 14$. However, since the patch size of the ViT encoder is $14$, we used as input spatial dimensions a direct multiple: $H = W = 252$ during training. As a consequence, during training $H_{\text{fire}} = W_{\text{fire}} = 25$. At test time, input and target dimensions are the same as for the CNN use case. To reduce overfitting issues, we used the LORA method \citep{hu2022lora} to fine-tune the ViT model with rank $r = 32$, $\alpha = 32$, and dropout $d_{\text{LORA}} = 0.1$. The channel dimension of the encoded feature map is $D_p = 384$, which is maintained after temporal encoding via ConvLSTM with kernel size $3 \times 3$.

The model extension for multi-modal data is done similarly to the CNN case with $D_{\text{env}} = 384$. This is to match the channel dimension of the final feature map. The same set of environmental predictors is used for this setting as for the CNN architecture above.

For the environmental-only architecture, the input and target spatial dimensions and processing are identical to the CNN use case. The temporal dimension differs as we use $T_{\text{env}} = 5$ for data augmentation. Both mid-resolution and low-resolution ViT-S encoders are randomly initialized and therefore do not use the LORA method for fine-tuning. In both encoders, for the position embedding, attention, and projection, we use a dropout rate $d_{\text{env}} = 0.2$ and a stochastic depth rate of $d_{\text{depth}} = 0.1$. For mid-resolution inputs, the patch size is $2$, and for low-resolution inputs, the patch size is $8$. The temporal encoding of the mid-resolution feature map is identical to the one used for the satellite image time series, and the temporal encoding of the low-resolution data is done through a one-layer LSTM with both input and output channel dimensions $D_{\text{low}} = 384$.

\subsection{CNN Training Parameters}

The CNN models are trained using the combined weighted cross-entropy and dice loss. The positive class (\textit{fire}) weight is $0.87$ and the negative class (\textit{no fire}) is $0.13$, found experimentally. Training is run over $20$ epochs with a batch size of $24$ samples on a NVIDIA GeForce RTX 3080 Ti GPU. The scheduler for the learning rate follows a $2$-epoch warm-up from the starting learning rate of $1e^{-7}$ to the base learning rate of $5e^{-6}$. Then the learning rate follows a cosine annealing of one cycle to the minimum learning rate of $1e^{-7}$ over the rest of the epochs. The optimizer used is ADAMW with a weight decay of $0.01$. During training, the augmentation pipeline first randomly crops the satellite input images to the training resolution, then resizes the images with a scale $s \in [0.9, 1]$. The images are randomly flipped horizontally and vertically, and Gaussian noise with variance $\sigma^2 \in [0.01, 0.1]$ is injected. Finally, we randomly sample the satellite image time series to extract $T = 5$ images (or pad when necessary). At test time, we center-crop the images to the required resolution and select the last $T = 5$ samples. For the multi-modal training, the non-spatial environmental data is not augmented, while for the environment-only architecture, we apply random horizontal and vertical flipping and Gaussian noise injection, similarly to the satellite image time series. The missing values in the environmental predictors, mainly caused by the NDVI and EVI as they are 16-day composites, are replaced during training with the value $0.0$.

\subsection{ViT Training Parameters}

Most of the ViT training parameters are the same as for the CNN models, except for the batch, which, despite also being $24$, is accumulated across two steps of $12$ for the ViT models. Moreover, during the training of the environmental-only use case, as we select $T_{\text{env}} = 5$ time steps, it is also necessary to randomly sample across the $8$ available samples. Finally, at test time across all modalities, we use the native temporal length for each sample, $8$ for the environmental data, and a variable length for the satellite image time series. The processing of the missing values for the environmental predictors is the same as for the CNN-based architecture.

\section{Ablation Study of the Impact of Satellite Image Time Series}
\label{sec:abl}

In Table \ref{tab:abl}, we analyze the performance of the multi-modal models in setting \Three with respect to the usage of time series. We compare our full multi-modal model using satellite image time series against a version using only the most recent image available before the prediction. In practice, for the CNN-based model, this impacts the number of parameters in the U-Net decoder as $D_i > D'_i$. 
Regardless of the architectures, the model performs best when presented with SITS rather than a single Sentinel-2 tile. As a consequence, we can hypothesize that dynamic factors directly linked to wildfire can be learned by the model from the temporal dimension of Sentinel-2.

\begin{table}[t]
\centering
\begin{tabular}{cc|cc}
\toprule
\textbf{Encoder} & \textbf{SITS} & \textbf{PRAUC} & \textbf{F1} \\
\midrule
ResNet-50 & \xmark & 42.4 & 48.3 \\
ResNet-50 & \cmark & \textbf{46.1} & \textbf{51.1} \\
\midrule
ViT-S & \xmark & 38.2 & 47.4 \\
ViT-S & \cmark & \textbf{43.9} & \textbf{50.0} \\
\bottomrule
\end{tabular}
\caption{Ablation study of SITS impact on the validation set performance.}
\label{tab:abl}
\end{table}

\section{Ignition Proxy Extension}
\label{sec:proxy}

We investigate the potential of including data about ignition proxies to better handle the Test Hard use case. We identify three different types of ignition proxies: yearly population density from WorldPop at 1 km resolution \citep{tatem2017worldpop}, static road density from GRIP4 at 8 km resolution \citep{meijer2018global}, and daily lightning density from WWLLN at 0.5 degree resolution \citep{kaplan2021wglc}. We extend the non-spatial representation of the environmental predictors for the multi-modal setup: \Three, with the ignition proxies, and train our best performing model for each architecture type: ResNet Multi-Modal, and ViT-S Multi-Modal. The results in Table \ref{tab:prox-res} show that the integration of ignition proxies only leads to a small improvement across both architectures. We thus argue that models would benefit from being trained on datasets considering hard negative sampling (similar to the way we sampled our Test Hard), where the modeling of ignition factors becomes the main discriminative factor.

\begin{table}[t]
    \centering
    \renewcommand{\arraystretch}{1.2}
    \resizebox{\textwidth}{!}{
    \begin{tabular}{ll|cccccc|cc}
        \toprule
        \multirow{2}{*}{\textbf{Encoder}} & \multirow{2}{*}{\textbf{Modality}} & \multicolumn{2}{c}{\textbf{Val}} & \multicolumn{2}{c}{\textbf{Test}} & \multicolumn{2}{c|}{\textbf{Test Hard}} & \multicolumn{2}{c}{\textbf{Avg}} \\
        \cmidrule(lr){3-4} \cmidrule(lr){5-6} \cmidrule(lr){7-8} \cmidrule(lr){9-10}
        & & PRAUC & F1 & PRAUC & F1 & PRAUC & F1 & PRAUC & F1 \\
        \midrule
        \multirow{2}{*}{ResNet-50} 
        & Multi-Modal & 46.1 & \underline{51.1} & \underline{57.0} & 60.3 & 27.1 & \underline{37.4} & \underline{43.4} & \underline{49.6} \\
        & Multi-Modal $\&$ Proxies & \textbf{46.7} & \underline{51.1} & \textbf{57.3} & \textbf{61.0} & \textbf{27.8} & \textbf{37.6} & \textbf{43.9} & \textbf{49.9} \\
        \midrule
        \multirow{2}{*}{ViT-S}
        & Multi-Modal & 43.9 & 50.0 & \textbf{56.3} & \textbf{59.2} & \underline{25.1} & \textbf{36.6} & \textbf{41.8} & \textbf{48.6} \\
        & Multi-Modal $\&$ Proxies & \textbf{44.5} & \textbf{50.9} & 55.3 & 56.7 & \textbf{25.4} & 35.2 & \underline{41.7} & 47.6 \\
        \bottomrule
    \end{tabular}}
    \caption{Performance comparison of Multi-Modal models with or without ignition proxy. \textbf{Bold} indicates the best metric value for each dataset split and model type, and \underline{underline} denotes a difference $\leq 0.5$ with the best option.}
    \label{tab:prox-res}
\end{table}

\section{Impact of Clouds}
\label{sec:cloud}

\begin{figure}[htpb]
    \makebox[\textwidth]{
    \resizebox{1.2\textwidth}{!}{
    \includegraphics[width=1.2\textwidth]{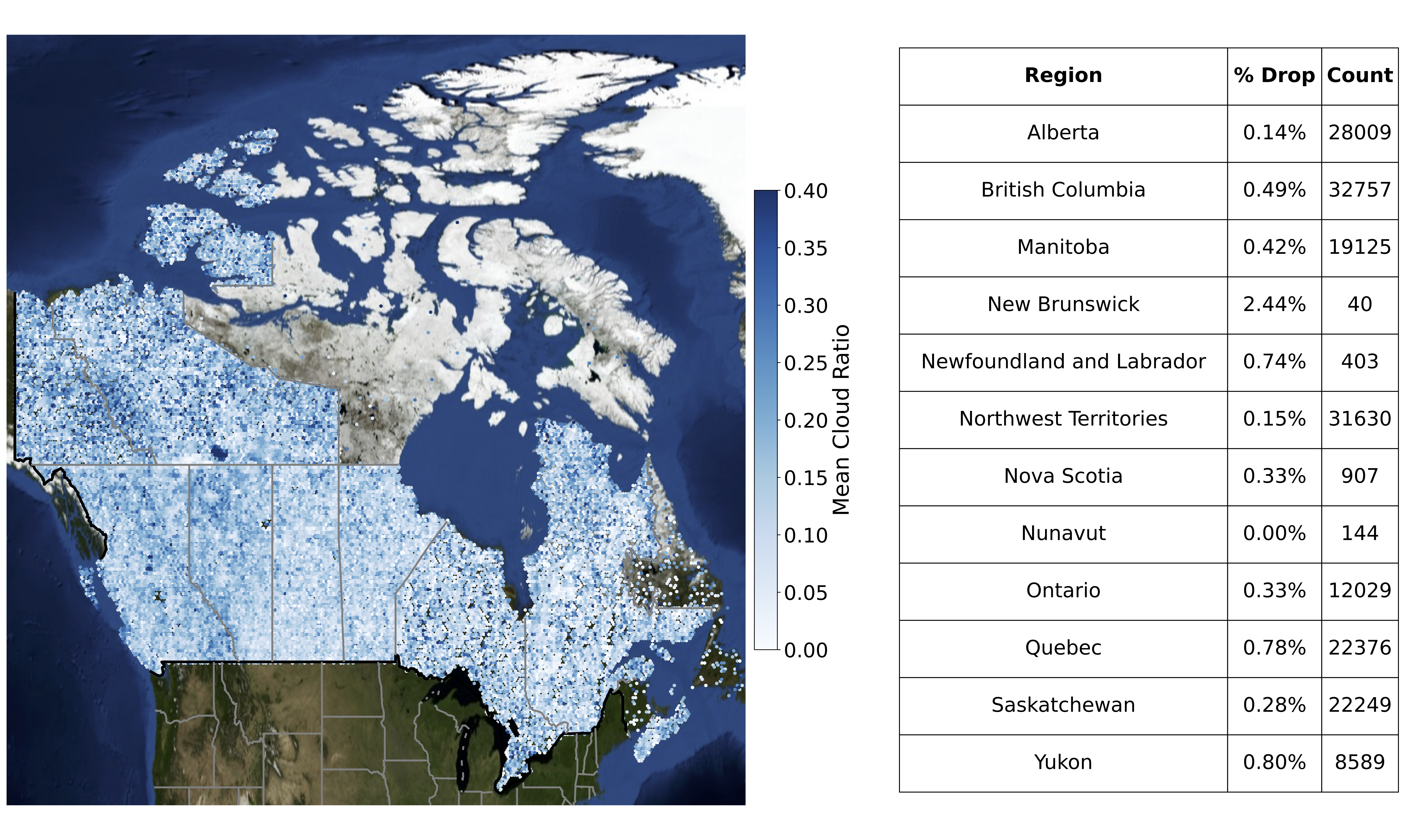}
    }
    }
    \caption{\textbf{(Left):} Average ratio of cloud pixels for the the Sentinel-2 images  following the initial cloud filtering. \textbf{(Right):} Statistics of the second cloud filter based on the ratio of cloud pixels: percentage of dropped samples, and post-filter count.}
    \label{fig:cloud}
\end{figure}

In CanadaFireSat, the impact of clouds is reduced by the temporal dimension of the data and by the usage of cloud-related bands. In Figure~\ref{fig:cloud}, we show that, despite an average cloud pixel ratio for the Sentinel-2 time series above $0.4$ for some samples, we only remove a small percentage of samples after the initial cloud filter, which is applied during the download of the images. In particular, we observe that Yukon, Northwest Territories, and British Columbia, with the Rocky Mountains, show a higher mean cloud ratio, while having a similar percentage of removed samples compared to the other provinces. This can be explained by the fact that we target a minimum of $3$ valid (cloud cover below $0.4$) S2 images across $40$ days for the whole S2 time series. As a consequence, models trained on CanadaFireSat are fairly robust to clouds thanks to the temporal dimension that enables training on valid S2 images only.

During training, as the time series can be incomplete due to cloud cover, additionally to the non-uniform temporal sampling of Sentinel-2, the model receives as input the day of the year (DOY) for each image in the SITS, therefore making it aware of the acquisition date, and more robust to the irregularly sampled S2 time series.

Finally, in \ref{sec:abl} we show that for potential extreme cases of cloud cover, where only one valid S2 image would be available prior to the fire occurring, both models (ResNet-50 and ViT-S) are robust to the absence of temporal information, as the performance reduction remains moderate (see Table \ref{tab:abl}).

\section{Test Hard Analysis}
\label{sec:app-test}

\begin{figure}[t!]
    \centering
    \includegraphics[width=0.8\textwidth]{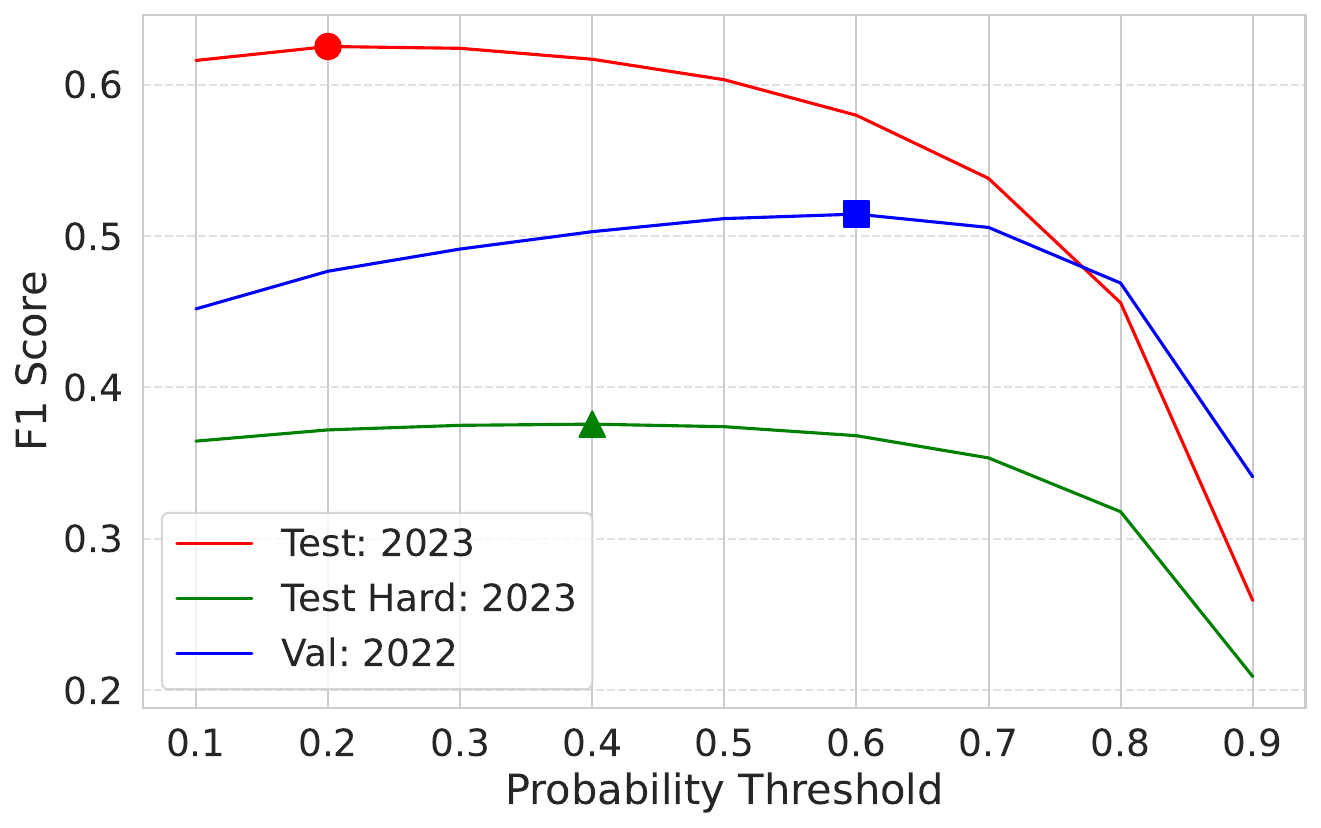}
    \caption{Analysis of the F1 score performance as a function of the probability threshold across all evaluation sets. The circle, square, and triangle represent the maximum value for each set.}
\label{fig:th}
\end{figure}

Figure \ref{fig:th} demonstrates the domain shift between the Val and the Test set, as the evolution of the F1 score with the probability threshold is centered around 0.5 for the Val set, presenting a normal behavior while being shifted towards a smaller threshold value for the Test set. As a consequence, the metrics in Table \ref{tab:results} might overestimate the model performance on the test set due to the extreme fire patterns during this year. For this purpose, we constructed the adversarial set named Test Hard for the year 2023 as described in Section \ref{sec:unbu}. Figure \ref{fig:th} also shows the delta in performance between Test and Test Hard: the centering of the maximum value for Test Hard is closer to the 0.5 threshold, representing a better alignment with the model behavior on the Val set.

\begin{figure}[t!]
 \captionsetup[subfigure]{justification=centering}
  \begin{subfigure}{\textwidth}
    \centering
    \includegraphics[width=0.8\textwidth]{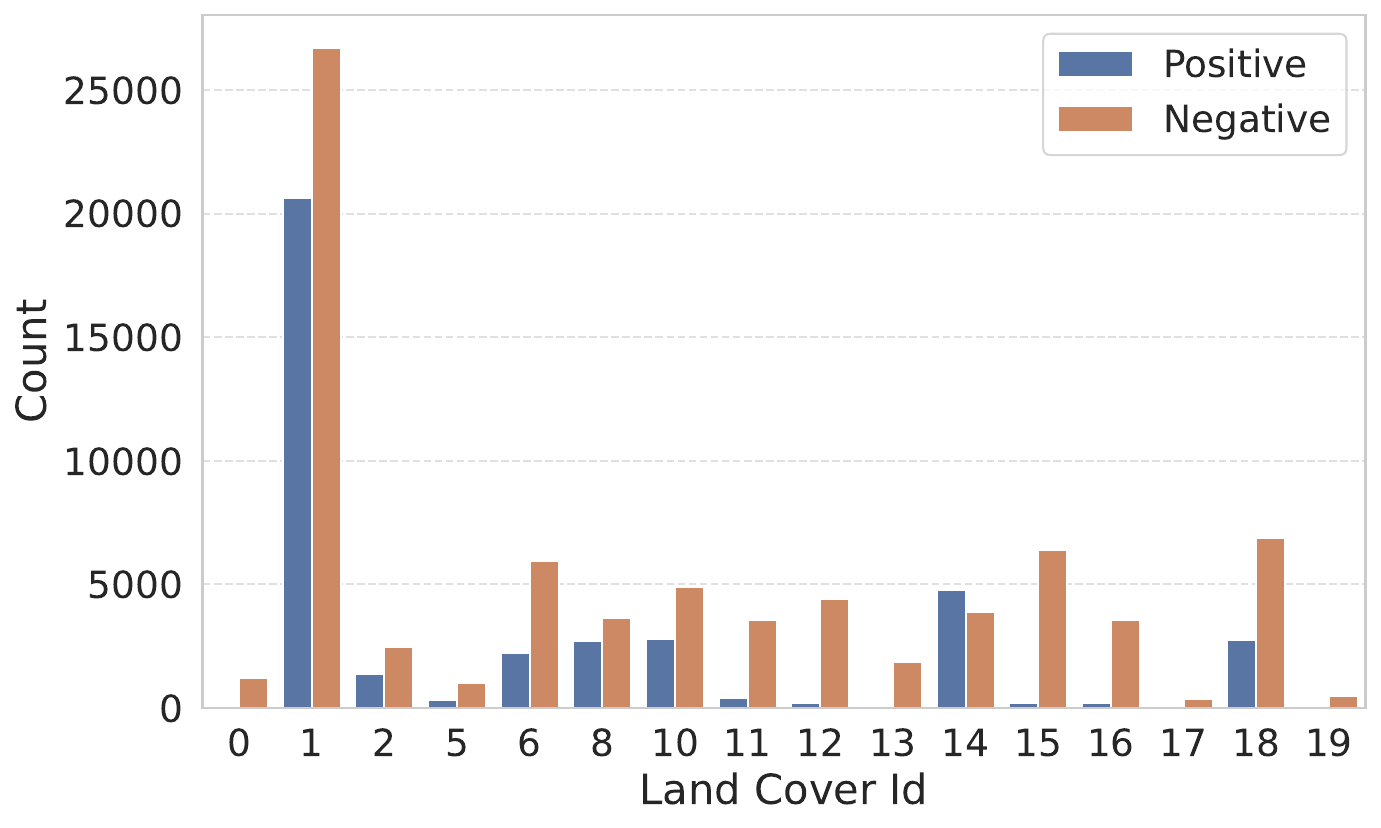}
  \subcaption{Land cover distribution for the Test set across positive and negative samples.}
  \label{fig:lc-dist}
  \end{subfigure}
  
  \vspace{0.5cm} 

  \begin{subfigure}{\textwidth}
    \centering
    \includegraphics[width=0.8\textwidth]{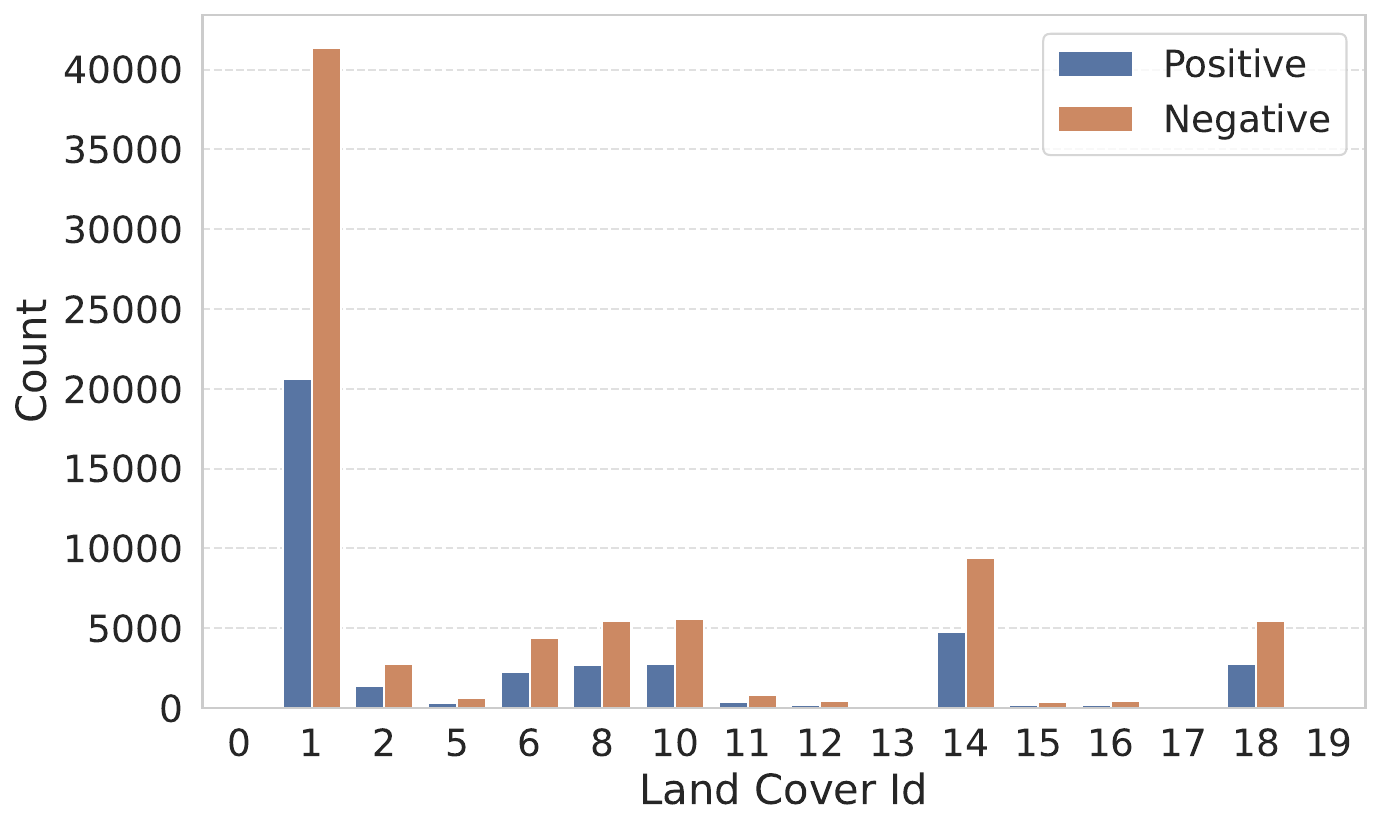}
  \subcaption{Land cover distribution for the Test Hard set across positive and negative samples.}
  \label{fig:lc-dist-hard}
  \end{subfigure}
  \caption{Comparison of the land cover distribution across the Test and Test Hard sets for the positive and negative samples.}
\label{fig:lc-dist-comp}
\end{figure}

Figure \ref{fig:lc-dist-comp} presents the change in land cover distribution for the negative samples between the two sets, Test and Test Hard, with respect to the positive samples. The stratification sampling done in Test Hard better aligns the categorical distributions for the negative and positive populations. 

\section{Deployment at Scale: Second Case Study}

\begin{figure}[htbp]
\centering
\resizebox{\columnwidth}{!}{
 \begin{tabular}{lc}
 \rotatebox{90}{\parbox{6cm}{\centering Original S-2 Tile}} & 
 \includegraphics[width=\textwidth, height=6cm]{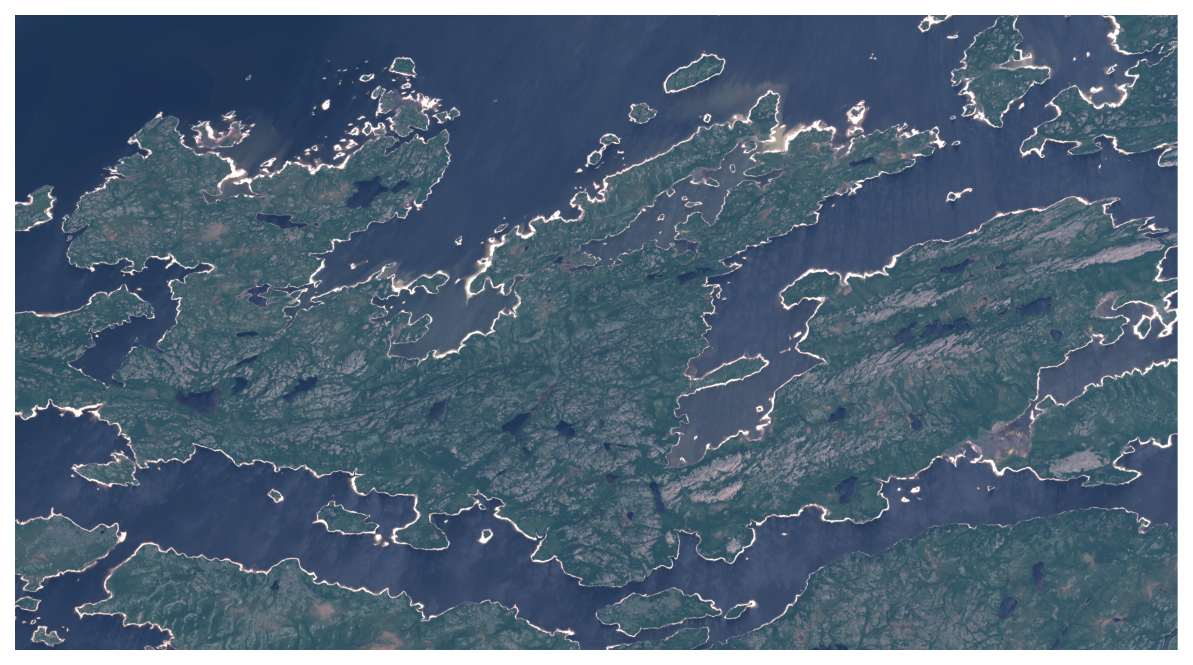} \\
 \rotatebox{90}{\parbox{6cm}{\centering Target Fire Polygon}} & 
 \includegraphics[width=\textwidth, height=6cm]{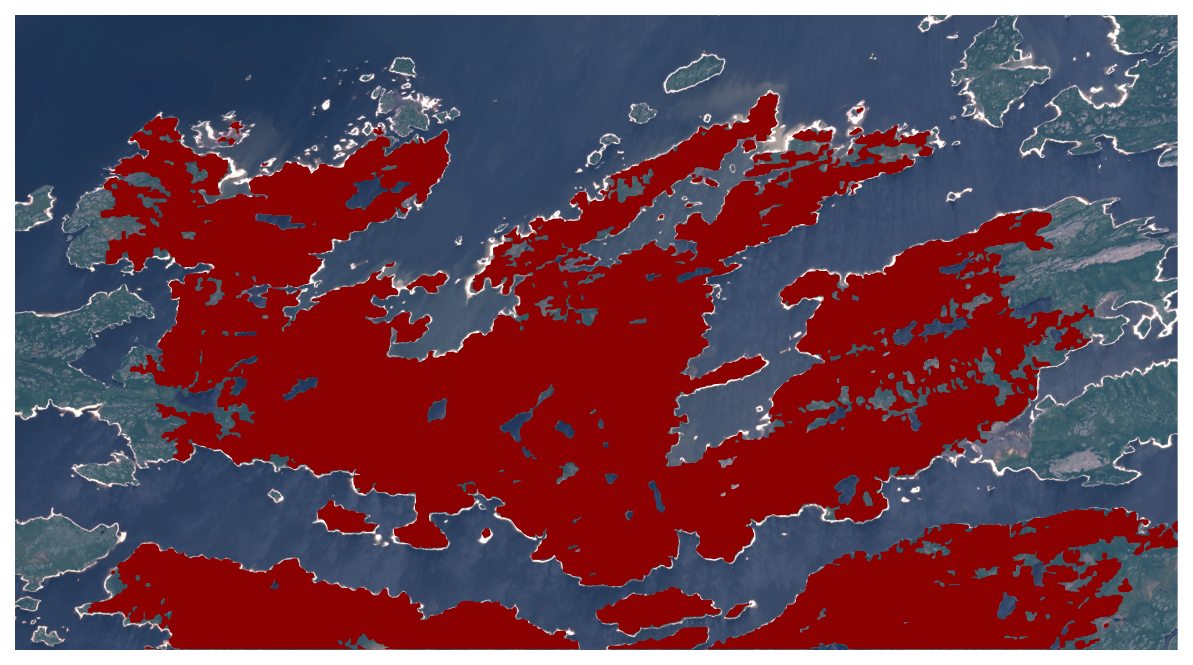} \\
 \rotatebox{90}{\parbox{6cm}{\centering Model Predictions}} & 
 \includegraphics[width=\textwidth, height=6cm]{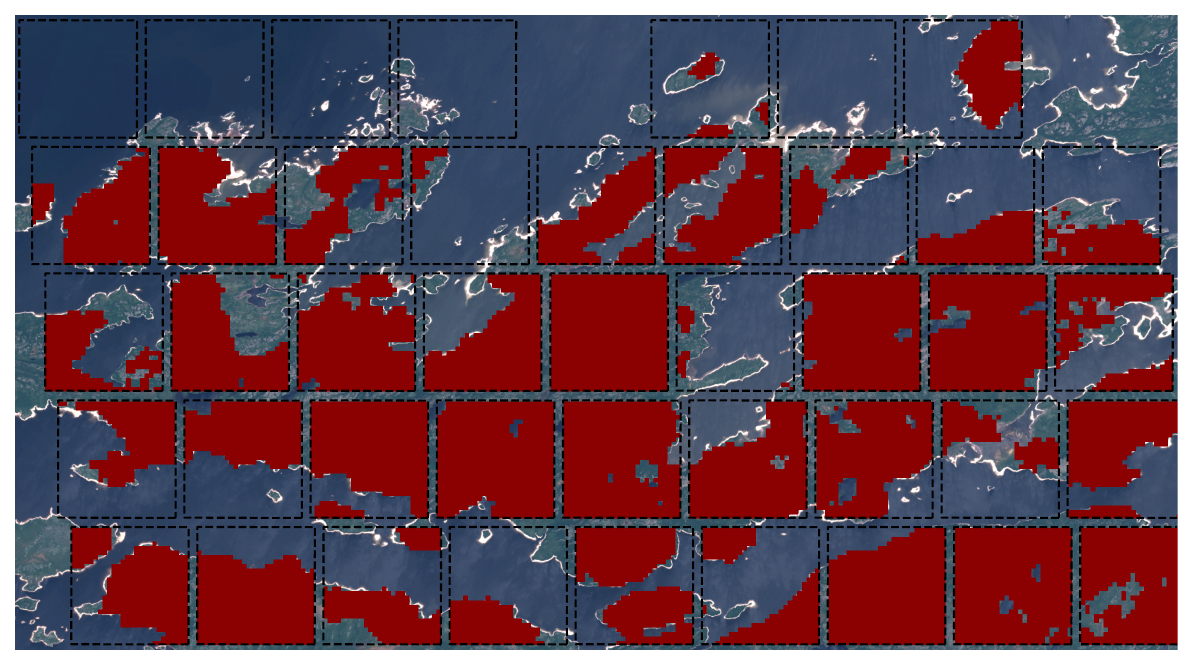}
 \end{tabular}}
 \caption{\textbf{Row 1} Sentinel-2 tile from $2023/06/28$ of size $14 \: \text{km} \times 26 \: \text{km}$ before a large wildfire in Québec. \textbf{Row 2} Fire polygons for the large wildfire on $2023/07/05$ over the same tile. \textbf{Row 3} Binary model predictions (in \textcolor{red}{\textbf{red}}) over the $2.64 \: \text{km} \times 2.64 \: \text{km}$ center-cropped positive samples (patches boundaries are outlined in \textbf{black}).}
 \label{fig:use case-bis}
\end{figure}

In Figure \ref{fig:use case-bis}, we present another case study for our CNN-based model in setting \Three. This example presents a large wildfire in Québec occurring on the $2023/07/05$, displayed over an RGB composite of a Sentinel-2 image of $14 \; \text{km} \times 26 \; \text{km}$. The predictions follow the same pattern as the actual wildfire, despite slightly overestimating its extent, as it can be seen on both sides of the Sentinel-2 tile, similarly to what we observed in Figure \ref{fig:use case}.

\end{document}